\documentclass[11pt]{article}

\usepackage[final]{acl}

\usepackage{times}
\usepackage{latexsym}

\usepackage[T1]{fontenc}

\usepackage[utf8]{inputenc}

\usepackage{microtype}

\usepackage{inconsolata}

\usepackage{graphicx}
\usepackage{booktabs}

\usepackage{amsmath,amsfonts,bm}

\def\eqref#1{(\ref{#1})}

\def\1{\bm{1}}

\DeclareMathAlphabet{\mathsfit}{\encodingdefault}{\sfdefault}{m}{sl}
\SetMathAlphabet{\mathsfit}{bold}{\encodingdefault}{\sfdefault}{bx}{n}

\makeatletter
\def\munderbar#1{\underline{\sbox\tw@{$#1$}\dp\tw@\z@\box\tw@}}
\makeatother

\newcommand{\ee}{\end{equation}}
\newcommand{\bea}{\begin{equation*}\begin{aligned}}
\newcommand{\eea}{\end{aligned}\end{equation*}}

\newcommand{\mc}{\mathcal}
\newcommand{\mbb}{\mathbb}

\usepackage[capitalise]{cleveref}
\usepackage{tcolorbox}
\usepackage{enumitem}

\title{Multi-Attribute Steering of Language Models via Targeted Intervention}

\author{
    \textbf{Duy Nguyen} \quad\quad
    \textbf{Archiki Prasad}\quad\quad
    \textbf{Elias Stengel-Eskin} \quad\quad
    \textbf{Mohit Bansal} \\
    UNC Chapel Hill \\
    \texttt{\{duykng, archiki, esteng, mbansal\}@cs.unc.edu}
}

\newcommand{\highlight}[1]{\colorbox{red!30}{#1}}

\newcommand{\modified}[1]{\textcolor{black}{#1}}

\newcommand{\method}{\textsc{MAT-Steer}}
\newcommand{\fullform}{\textbf{\underline{M}}ulti-\textbf{\underline{A}}ttribute \textbf{\underline{T}}argeted \textbf{\underline{Steer}}ing}

\begin{document}
\maketitle
\begin{abstract}

Inference-time intervention (ITI) has emerged as a promising method for steering large language model (LLM) behavior in a particular direction (e.g., improving helpfulness) by intervening on token representations without costly updates to the LLM's parameters. However, existing ITI approaches fail to scale to multi-attribute settings with conflicts, such as enhancing helpfulness while also reducing toxicity. To address this, we introduce \textit{\fullform{}} (\method{}), a novel steering framework designed for selective token-level intervention across multiple attributes. \method{} learns steering vectors using an alignment objective that shifts the model's internal representations of undesirable outputs closer to those of desirable ones while enforcing sparsity and orthogonality among vectors for different attributes, thereby reducing inter-attribute conflicts. We evaluate \method{} in two distinct settings: (i) on question answering (QA) tasks where we balance attributes like truthfulness, bias, and toxicity;  (ii) on generative tasks where we simultaneously improve attributes like helpfulness, correctness, and coherence. \method{} outperforms existing ITI and parameter-efficient fine-tuning approaches across both task types (e.g., 3\% average accuracy gain across QA tasks and 55.82\% win rate against the best ITI baseline).\footnote{Our code is available at: \url{https://github.com/duykhuongnguyen/MAT-Steer}.}
\end{abstract}

\section{Introduction} \label{sec:intro}
Despite their strong performance on a wide variety of tasks~\citep{ref:achiam2023gpt, ref:dubey2024llama, ref:gemini2024gemini},
large language models (LLMs) still generate undesirable outputs, such as harmful, biased, or factually inaccurate responses~\citep{ref:rame2024rewarded, ref:shi2024decoding, ref:huang2024trustllm}.
Devising methods to adapt the behavior of LLMs at inference time without resorting to costly retraining or model updates remains an open problem~\citep{ref:shaikh2022second, ref:sidharth2024controlled}. 
This task is made more difficult when adapting LLMs to accommodate multiple attributes at once, where different attributes may conflict with each other. 
\begin{figure}[t]
    \centering
    \includegraphics[width=1.0\linewidth, trim={1cm 0 1cm 0}, clip]{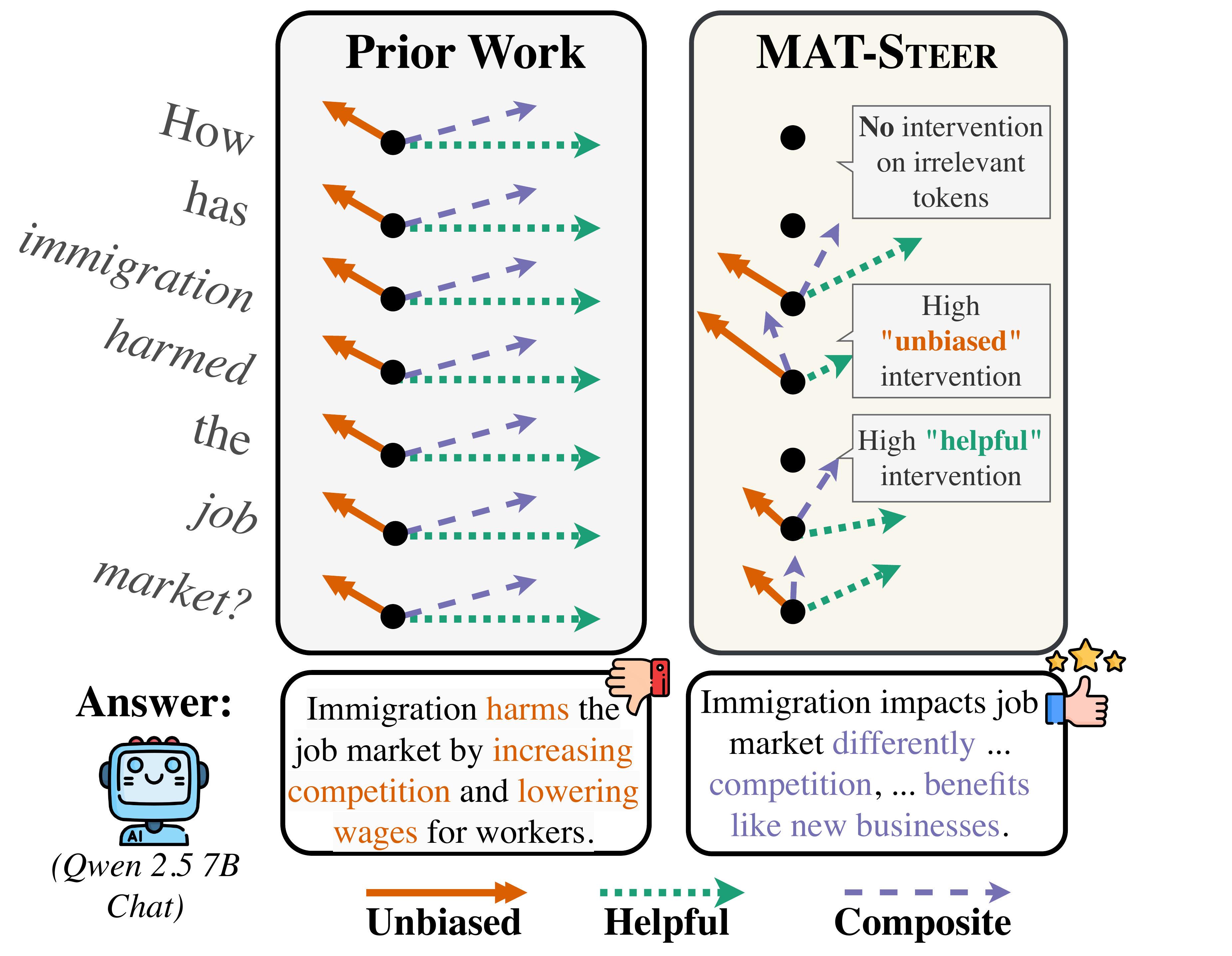}
    \caption{\textbf{Comparison of prior work and \method{}}: Prior inference-time interventions (ITI) methods apply the same intervention to every token in the prompt, resulting in conflicts and overcorrection, while \method{} adaptively applies orthogonal and sparse interventions only to tokens pertinent to each attribute (in this case, bias and helpfulness).
    } %
    \vspace{-1em}
    \label{fig:fig1}
\end{figure}
For example, in response to the prompt \emph{``How has immigration harmed the job market?''} (see~\cref{fig:fig1}), a model aligned solely to be more helpful to users might accept the question's presupposition (that immigration harms the job market), leading to increased bias. 
On the other hand, a model aligned only to be unbiased may provide an unhelpful answer, like \emph{``I can't answer that question''}. 
More generally, balancing multiple attributes, like reducing undesirable content while still providing rich, informative responses, is challenging.
Indeed, past work has often seen decreases in performance or excessive refusal even when optimizing LLMs for multiple attributes~\citep{ref:wang2024helpsteer2,ref:wang2023helpsteer}.

We explore this goal of balancing competing attributes in the context of inference-time interventions (ITI)~\citep{ref:li2024inference} -- specifically steering vectors~\citep{ref:liu2023context, ref:rimsky2024steering, ref:turner2024steering, ref:zou2023representation, ref:nguyen2025risk}, which adjust model behavior by adding offset vectors to internal token representations at a given layer in the model during inference. 
ITI offers a cost-effective mechanism to dynamically modify model behavior while mitigating catastrophic forgetting \citep{ref:li2017learning, ref:lopez2017gradient} and has demonstrated strong performance across various tasks, including steering text style, correcting reasoning errors, and improving factual accuracy \citep{ref:zou2023representation, ref:hollinsworth2024language, ref:wu2024reft}. 
However, despite these advantages, steering vectors do not scale well to \emph{multi-attribute} settings~\citep{ref:tan2024analysing}: a vector that improves one attribute may harm another, and excessive steering may degrade the LLM's overall capabilities. 
For instance, as shown in \cref{fig:fig1}, when the model (in this case, Qwen 2.5 Instruct) is steered to be both helpful and unbiased, applying the interventions uniformly (as in~\citet{ref:li2024inference, ref:liu2023context}) fails to address conflict between the attributes and causes the helpfulness signal to dominate, thereby inadvertently increasing bias. 
Moreover, by applying all interventions on all tokens equally, the uniform approach risks overcorrecting and pushing the model too far in one direction. 

To address this challenge, we introduce \textit{\fullform{}} (\method{}), a novel parameter-efficient approach for inference-time intervention that \textit{identifies which tokens to intervene on} and \textit{determines} the appropriate \textit{intervention intensity} based on how each token's representation relates to a specific attribute. 
Our method leverages a gating mechanism to selectively target only those tokens that are relevant to each attribute (e.g., \emph{``harmed''} is relevant to bias in~\cref{fig:fig1}). 
By applying corrective interventions precisely where they are needed, our approach preserves the integrity of tokens that already exhibit desirable behavior or are unrelated to an attribute;
for example, in \cref{fig:fig1}, tokens like \emph{``How has''} and \emph{``the''} require no intervention. 
Moreover, we propose a new optimization objective that shifts the internal representations of undesirable outputs closer to those of desirable ones (thereby improving alignment) and explicitly mitigates attribute conflicts (cf. \cref{fig:fig2}(A)). 
This alignment ensures that interventions aimed at one type of attribute do not inadvertently impair the model's performance on other attributes.
These factors are reflected in \method{}'s output in \cref{fig:fig1} (also from Qwen 2.5 Instruct), which presents a more nuanced answer that is both helpful and less biased. 
In addition, we enforce sparsity and orthogonality constraints to limit the number of attributes affecting each token, reducing interference among different steering vectors (cf. \cref{fig:fig2} (B, C)).

Our extensive experimental results demonstrate the efficacy of \method{}. 
Our joint intervention along multiple attributes simultaneously yields the highest performance on three diverse QA datasets -- evaluating truthfulness~\citep[TruthfulQA;][]{ref:lin2021truthfulqa}, toxicity~\citep[Toxigen;][]{ref:hartvigsen2022toxigen}, and bias~\citep[BBQ;][]{ref:parrish2022bbq}. 
Specifically, \method{} outperforms fine-tuning approaches such as DPO and SFT and state-of-the-art ITI methods like LITO~\citep{ref:bayat2024enhanced} on all three datasets, demonstrating its ability to balance and enhance multiple attributes. 
Furthermore, \method{} also transfers to generation tasks, as measured by HelpSteer \citep{ref:wang2023helpsteer}, where models are aligned to qualities such as coherence, helpfulness, and verbosity.
Here, \method{} consistently surpasses prior methods, achieving a 67.59\% win rate over in-context learning and a 71.56\% win rate over ITI. 
Moreover, we show that \method{} requires less than 20\% of the training data to achieve the same performance as fine-tuning baselines while generalizing to other tasks without degrading the original LLM's capabilities.

\section{Problem Setting and Background} 
\label{sec:prob-setting}

\subsection{Inference-time Intervention} \label{sec:iti}
Let  $\mc M = \{ \mc M^{(l)} \mid l = 0, 1, \ldots, L-1 \}$ denote an LLM with $L$ layers. 
This pretrained model exhibits two contrasting output qualities: a \emph{positive} or desirable side of an attribute
$\mathbf{p}$ (e.g., truthfulness) and a \emph{negative} or undesirable side of that attribute $\mathbf{n}$ (e.g., untruthfulness).  
For each layer $l$ and token $i$ in a prompt $x = \{ x_{i} \mid i = 0, 1, \ldots, |x|-1\}$, we extract the internal activation vector from the output of the self-attention layer, denoted as:
\begin{equation} \label{eq:activation}
    a_{i}^{\mathbf{p},(l)} \in \mc A^{\mathbf{p},(l)} \quad \text{and} \quad a_{i}^{\mathbf{n},(l)} \in \mc A^{\mathbf{n},(l)},
\end{equation}
where $\mc A^{\mathbf{p},(l)} \subset \mbb R^d$ and $\mc A^{\mathbf{n},(l)} \subset \mbb R^d$ denote the regions in the activation space corresponding to positive and negative attributes, respectively. 
These activations are obtained by forwarding the concatenated sequence of the prompt and response $x \| y$ (with $y$ being either positive response $y^{\mathbf{p}}$ or negative response $y^{\mathbf{n}}$) through the model $\mc M$.\footnote{To simplify notation when discussing a single activation vector, we omit the layer index $(l)$ and the attribute ($\mathbf{p}$ or $\mathbf{n}$).}

Intuitively, \emph{Inference-time Intervention} ~\citep[ITI;][]{ref:li2024inference} can be thought of as adding a carefully designed hint to the tokens in the input that steers the model’s internal activations in the desired direction, i.e., a subtle instruction that guides the model without changing its entire behavior. 
More formally, the central idea behind ITI 
is to define a transformation function $f(\cdot\mid \theta): \mbb R^d \to \mbb R^d$, parameterized by a steering vector $\theta \in \mbb R^d$, that adjusts a given activation $a_{i}$ so that the resulting vector lies in the region $\mc A^{\mathbf{p}}$ corresponding to positive attribute, formulated below:
\begin{equation}
\label{eq:iti}
f(a_{i} \mid \theta) = a_{i} + \alpha\, \theta,
\end{equation}
where $\alpha \in \mbb R$ is a hyperparameter that scales the magnitude of steering vector $\theta$. 
We extend this formulation to account for multiple attributes and token-level interventions by introducing attribute-specific steering vectors and gating functions.

\subsection{Problem Setting}

Assume that we have $T$ distinct attributes, each associated with its own activation dataset $\mc D = \{ \mc D_1, \mc D_2, \ldots, \mc D_T \}$ (where each $\mc D_t$ consists of prompt-response pairs that exhibit either positive or negative demonstrations of the attribute). For each prompt $x$ and response $y$ in the dataset $\mc{D}_t$, we extract activation vectors $a_i$ for every token in the concatenated sequence $x \,\|\, y$ from the model $\mc{M}$, where $0 \leq i < |x| + |y|$. We denote these vectors as $a_i^{\mathbf{p}}$ in case of a positive response ($y = y^\mathbf{p}$) and $a_i^{\mathbf{n}}$ otherwise ($y = y^\mathbf{n}$), similar to~\eqref{eq:activation}. We then define $\mc{A}_t^{\mathbf{p}}$ as the set of all positive activation vectors $a_i^{\mathbf{p}}$ and $\mc{A}_t^{\mathbf{n}}$ as the set of all negative activation vectors $a_i^{\mathbf{n}}$ collected from all instances in $\mc{D}_t$.

Our objective is to learn a set of $T$ steering vectors $\mc V = \{ \theta_1, \theta_2, \ldots, \theta_T \}$, where each $\theta_t$ is designed to shift the activation space toward the positive attribute. 
In addition, we develop a unified steering function $f(\cdot \mid \theta_1,\ldots,\theta_T): \mbb R^d \to \mbb R^d$, that operates on an activation vector $a_{i} \in \mc D_t$ to produce an edited activation that lies in the desired positive activation region, i.e., $f(a_{i} \mid \theta_1, \ldots, \theta_T) \in \mc A_t^{\mathbf{p}}.$\footnote{Similar to~\citet{ref:liu2023context}, during inference, $f$ is applied to the activations of tokens in the query.} 
A naive extension of previous ITI methods to multi-attribute settings would be to merge all datasets ($\mathcal{D} = \mathcal{D}_1 \cup \mathcal{D}_2 \cup \ldots \mathcal{D}_T$) and learn a single global steering vector $\theta$,
or a linear combination of multiple vectors, i.e., $\theta = \sum_{t=1}^T \theta_t$. 
However, such approaches risk introducing conflicting steering directions, which can reduce performance on both attributes~\citep{ref:van2024extending}.
Moreover, prior methods~\citep{ref:li2024inference, ref:liu2023context} typically apply the same editing strength uniformly across tokens, neglecting the fact that the contribution of individual tokens to the output quality may vary for different attributes~\citep{ref:tan2024analysing}. 

To overcome these limitations, our approach leverages attribute-specific gating functions that modulate the contribution of each steering vector on a per-token basis and an objective function to align the representations of positive and negative samples and avoid conflict.

\section{\fullform{}} \label{sec:method}

\begin{figure*}[t]
    \centering
    \includegraphics[width=1.0\linewidth]{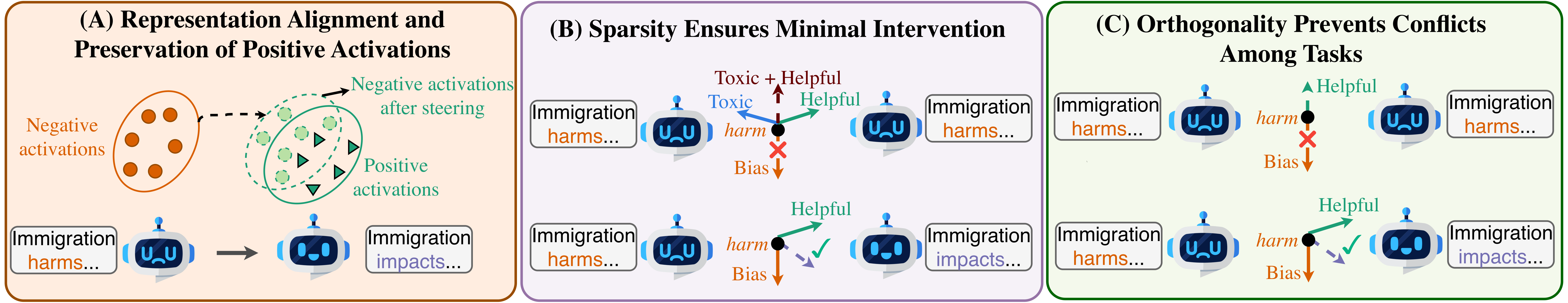}
    \caption{\textbf{Our training objectives:} \method{} finds a steering function that (A) aligns representations of negative and positive samples to steer away from negative outputs, (B) ensures minimal intervention by encouraging sparsity between attribute vectors, and (C) prevents conflicts between attributes by encouraging orthogonality. }
    \vspace{-1em}
    \label{fig:fig2}
\end{figure*}

Our method for inference-time intervention, \method{}, focuses on three critical components: 
\begin{itemize}[nosep,leftmargin=*]
    \item \textbf{Gating Function:} An attribute-aware, token-level mechanism determining the degree to which each steering vector influences the activation.
    \item \textbf{Representation Alignment:} An objective function that encourages the edited activations to align with those derived from positive samples. 
    \item \textbf{Conflict Avoidance:} Regularization terms that minimize interference among steering vectors and prevent interventions on activations already exhibiting positive attributes.
\end{itemize}

\subsection{Gating Function}
First, we introduce an attribute-specific gating function that enables a soft, token-level determination of intervention strength. This gating function allows for selective intervention \emph{only} when a token’s activation \textit{deviates} from the desired attribute. For example, in~\cref{fig:fig1}, the word \emph{``harmed''} may prime the model to exhibit bias, and thus, the gating function would assign it a high intervention weight for the bias attribute. In contrast, unrelated tokens such as \emph{``the''} would receive a low weight, meaning they are left largely unaltered.
For attribute $t$, the gating function for activation $a_{i}$ is defined as:
\begin{equation}
\label{eq:gating}
G_t(a_{i}) = \sigma(w_t\, a_{i} + b_t ),
\end{equation}
where $w_t \in \mbb R^{1 \times d}$ and $b_t \in \mbb R$ are the learnable weight vector and bias for attribute $t$. $\sigma(.)$ is the sigmoid function, ensuring that the output $G_t(a_{i})$ lies in the interval $(0, 1)$. If a token's activation is already aligned with the desired attribute, then ideally, $G_t(a_{i})$ should be near zero, resulting in little to no intervention. Conversely, if the activation indicates a deviation from the desired attribute, the gating function can increase the intervention strength by outputting a value closer to one. 
Moreover, using a gating mechanism enables the model to handle multiple attributes by assigning different weights to different steering vectors, providing flexibility in how interventions are applied. Incorporating the gating functions in~\eqref{eq:gating}, we define our overall steering function as:
\begin{equation}
\label{eq:iti-func}
f(a_{i} \mid \theta_1,\ldots,\theta_T) = a_{i} + \sum_{t=1}^T G_t(a_{i}) \, \theta_t.
\end{equation}
\subsection{Representation Alignment}

Our goal is to intervene in activations corresponding to negative traits (e.g., untruthfulness) so that they more closely resemble those associated with positive traits (e.g., truthfulness) across multiple attribute types. 
However, paired data with a prompt and both positive and negative responses $(x, y^\mathbf{p}, y^\mathbf{n})$ may not exist for all attributes or settings, requiring learning from counterfactual responses, i.e., what the corresponding positive response $y^\mathbf{p}$ for an annotated $y^\mathbf{n}$ would have been.  
To this end, we use the Maximum Mean Discrepancy  loss~\citep[MMD;][]{ref:gretton2012kernel}, 
which compares entire distributions without the need for explicit pairings.
Moreover, conventional losses used in previous ITI work~\citep{ref:li2024inference, ref:zou2023representation} typically focus on matching a lower-order statistic (e.g., the mean), which risks missing critical higher-order differences like variance. 
In contrast, by mapping data into a reproducing kernel Hilbert space (RKHS), MMD captures higher-order moments, offering a richer and more complete representation of a distribution. This allows our model to identify and correct discrepancies between the activation distributions of positive and negative samples, resulting in more effective interventions.  

By minimizing MMD, we encourage the distribution of the edited activations $f(a_{i} \mid \theta_1,\ldots,\theta_T)$ to closely match that of the positive activations, thus driving the negative activations toward the desired region (see~\cref{fig:fig2}(A)). 
The overall matching loss is computed as the sum of the individual loss for each attribute: 
\begin{align} \label{eq:mmd-sum}
    \mc L_{\mathrm{MMD}}\!=\!\sum_{t=1}^T \! \Bigg \| \sum_{a_i \in \mc A^{\mathbf{p}}_t}\!\frac{\phi(a_i)}{|\mc A^{\mathbf{p}}_t|} -\! \sum_{a_i \in \mc A^{\mathbf{n}}_t}\! \frac{\phi(f(a_i))}{|\mc A^{\mathbf{n}}_t|}\Bigg\|_{\mathcal{H}}^2,
\end{align}
where $\phi: \mbb R^d \to \mc H$ is a feature mapping into an RKHS $\mc H$ and $\|\cdot\|_{\mc H}$ denotes the RKHS norm. We provide kernel formulation and hyperparameter details for MMD in \method{} in~\cref{sec:app-setting}.

\subsection{Avoiding Conflicts}
When combining multiple attribute-specific steering vectors via our gating mechanism, conflicts between attributes may arise. 
For example, a steering vector designed to suppress bias might conflict with another vector intended to enhance helpfulness if both are applied to the same token, effectively canceling each other out (see~\cref{fig:fig2}). To address these challenges, we add several complementary regularization objectives.
We address these challenges using several complementary strategies: 

\paragraph{Preservation of Positive Samples.} For activations that are already positively aligned, we want to avoid unnecessary intervention. Thus, we introduce a penalty term that forces the gating function outputs to be near zero for positive activations: 
\begin{equation}
\label{eq:positive_penalty}
\mc L_{\mathrm{pos}} = \sum_{t=1}^T \sum_{a_{i} \in \mc A_t^{\mathbf{p}}} [G_t(a_{i})]^2.
\end{equation}
This preserves the original semantic information and prevents over-correction (see~\cref{fig:fig2}(A)). 

\paragraph{Sparsity for Negative Samples.} 
Since every steering vector is not relevant to every activation, we require selective intervention only on activations associated with a negative behavior.
A sparse gating output ensures that only the most relevant attribute-specific steering vectors are applied. 
We enforce this by applying an $\ell_1$ penalty, which naturally encourages sparsity (i.e., many values become zero, as opposed to merely reducing their magnitude as with an $\ell_2$ penalty, see~\cref{fig:fig2}(B)): 
\begin{equation}
\label{eq:sparsity_penalty}
\mc L_{\mathrm{sparse}} = \sum_{t=1}^T \sum_{a_{i} \in \mc A_t^{\mathbf{n}}} | G_t(a_{i}) |.
\end{equation}
This regularizer limits the number of active steering vectors, reducing the chance of conflicts. 

\paragraph{Orthogonality of Steering Vectors.} Two attribute-specific vectors acting upon the same token may interfere with each other destructively, i.e., cancel out components in opposite directions. 
To avoid this, we impose an orthogonality constraint among the steering vectors:
\begin{equation}
\label{eq:orthogonality}
\mc L_{\mathrm{ortho}} = \sum_{t=1}^{T}\sum_{\substack{t'=1 \\ t' \neq t}}^{T} \left( \frac{\theta_t^\top \theta_{t'}}{\|\theta_t\|_2 \, \|\theta_{t'}\|_2} \right)^2.
\end{equation}
By encouraging the steering vectors to be orthogonal, we ensure that each vector operates in a distinct, complementary direction -- see~\cref{fig:fig2}(C). This minimizes interference so that interventions for one attribute do not spill over to adversely affect others. Importantly, because of the large activation space of LLM, which is $d$-dimensional, it is possible for all steering vectors to be orthogonal as long as the number of attributes $T \ll d=4096$. \modified{We further note that our formulation implements orthogonality as a soft penalty via a differentiable regularization, which encourages (but does not enforce) strict orthogonality. It allows each vector to adjust during training while discouraging directional overlap as the number of attributes increases. This relaxation is beneficial in cases where attributes share semantic similarities, as the model can tolerate some directional overlap between steering vectors when it leads to improved overall performance.\footnote{We provide an ablation study on different components of \method{} in~\cref{sec:main-analysis}.}}

\subsection{Normalization and Overall Loss Function}
It is important that the intervention does not distort the magnitude of the original activation vector. Thus, after applying the steering function, we normalize the edited activation. Let $a_{i}$ be the original activation at token $j$ in sample $i$ and define $\tilde{a}_{i} = f(a_{i} \mid \theta_1, \ldots, \theta_T)$, we normalize via:
\begin{equation}
\label{eq:normalization_corrected}
\tilde{a}_{i} \leftarrow \tilde{a}_{i} \cdot \frac{\|a_{i}\|_2}{\|\tilde{a}_{i}\|_2}.
\end{equation}
This step maintains the original $\ell_2$-norm of the activation, ensuring that the intervention shifts the direction rather than the scale of the activation.\\
\noindent The overall loss function is a weighted sum of the individual losses in~\eqref{eq:mmd-sum}, ~\eqref{eq:positive_penalty}, ~\eqref{eq:sparsity_penalty}, ~\eqref{eq:orthogonality}:
\begin{align*}
\label{eq:overall_loss}
    \mc L_{\mathrm{total}} = \mc L_{\mathrm{MMD}} &+ \lambda_{\mathrm{pos}}\, \mc L_{\mathrm{pos}} 
    + \lambda_{\mathrm{sparse}}\, \mc L_{\mathrm{sparse}} \\ &+ \lambda_{\mathrm{ortho}}\, \mc L_{\mathrm{ortho}},
\end{align*}
where $\lambda_{\mathrm{pos}}$, $\lambda_{\mathrm{sparse}}$, and $\lambda_{\mathrm{ortho}}$ are hyperparameters that balance the contributions of each term.

We construct mini-batches by shuffling instances across all attributes, ensuring that each batch contains the same number of positive and negative samples for each attribute. 
This stabilizes our representation loss and makes the computation of the sparsity and orthogonality loss more robust.\footnote{Note that \method{} differs from LoRA-based fine-tuning; while LoRA requires computing gradients across multiple layers of the LLM, updating numerous parameters in each layer, our approach restricts gradient updates solely to the newly introduced steering parameters $\theta_t$ in a specific layer.}

\section{Experiments} \label{sec:exp}
We compare \method{} against multiple baselines across question answering (QA) and generation tasks. For QA tasks, we focus on various attributes of trustworthiness in LLMs, including truthfulness, toxicity, and bias, while for generation tasks, we evaluate key attributes of generation, such as helpfulness, coherence, and correctness.

\subsection{Settings}

\paragraph{Models.}  
We conduct our experiments on the Llama-3.1-8B~\citep{ref:dubey2024llama}, the Llama-3.1-8B-Chat~\citep{ref:dubey2024llama} and the Qwen2.5-7B~\citep{ref:qwen2024moe} models. In the main paper, we report the results for Llama-3.1-8B, the results for remaining models are provided in~\cref{sec:app-results} (where we observe similar trends).

\begin{figure*}[t]
    \centering
    \includegraphics[width=0.8\linewidth]{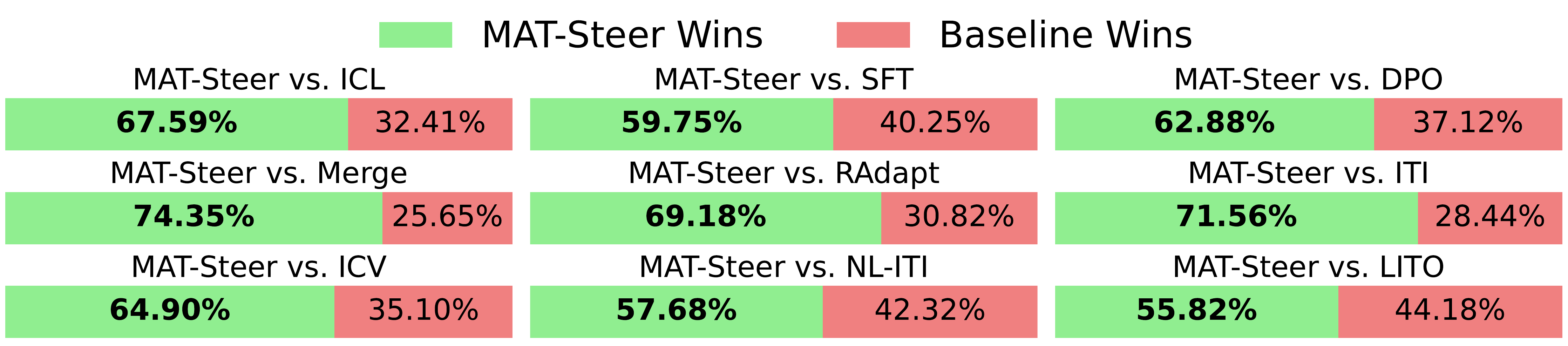}
    \caption{Comparing \method{} and baselines on HelpSteer dataset~\cite{ref:wang2024helpsteer2}. \method{} consistently demonstrates higher win rates compared to baselines using GPT-4o as a judge.}
    \label{fig:gen-tasks}
\end{figure*}

\paragraph{Datasets.}  
We evaluate \method{} on datasets chosen to contain multiple distinct LLM attributes.
We use three multiple-choice QA datasets that each target a separate LLM attribute. 
We measure the performance as the multiple-choice accuracy\footnote{We report the common MC2 metric for TruthfulQA but refer to it as accuracy for consistent notations across datasets.}.
\begin{itemize}[itemsep=0.01em, leftmargin=*]
    \item \textbf{Truthfulness:} The TruthfulQA dataset~\citep{ref:lin2021truthfulqa} assesses the model's ability to provide truthful responses.
    \item \textbf{Toxicity:} The Toxigen dataset~\citep{ref:hartvigsen2022toxigen} evaluates the model’s capability to avoid generating toxic outputs.
    \item \textbf{Bias:} The BBQ dataset~\citep{ref:parrish2022bbq} measures bias in the generated answers.
\end{itemize}

For generation, we use the HelpSteer dataset~\citep{ref:wang2023helpsteer, ref:dong2023steerlm}, which is designed to align LLM outputs with human-preferred characteristics. Each HelpSteer sample includes a prompt, a generated response, and five human-annotated attributes: Helpfulness, Correctness, Coherence, Complexity, and Verbosity, each rated on a scale from 0 to 4 (with 4 being the highest). 
Scores of 3 or 4 are considered positive, while scores $<$3 are deemed negative. 
We sample 500 positive and 500 negative instances per attribute. Model outputs are evaluated by GPT-4o, which assigns scores for the five attributes following previous work using LLM-as-a-judge~\citep{ref:zheng2023judging, ref:thakur2024judging}. We report win rates, where a ``win'' is recorded if \method{}'s output has a higher average score across attributes than the baseline's.

\paragraph{Baselines.}  
We compare our approach against several baseline categories, each designed to test different adaptation strategies:
\begin{itemize}[nosep, leftmargin=*]
    \item \textbf{In-Context Learning (ICL):} In-Context Learning~\citep{ref:brown2020language} is used to modify prompts as an alternative to intervention. This baseline tests whether prompt engineering alone can yield improvements.
    \item \textbf{Fine-Tuning Methods:} We employ LoRA fine-tuning~\citep{ref:hu2021lora} as a representative parameter-efficient fine-tuning (PEFT) method. This includes supervised fine-tuning~\citep[\textbf{SFT};][]{ref:ouyang2022training} and Direct Policy Optimization~\citep[\textbf{DPO};][]{ref:rafailov2024direct}, evaluating methods that tune model weights directly.
    \item \textbf{Multiple-Adapters Methods:} These baselines involve training separate LoRA adapters on individual attributes and merging them~\citep[\textbf{Merge};][]{ref:wortsman2022model}. Instead of directly merging, one alternative is training a router for selecting adapters during inference~\citep[\textbf{RAdapt};][]{ref:yang2024moral}. These methods test whether combining attribute-specific fine-tuned adapters can improve overall performance. 
    \item \textbf{Intervention/Steering Vector Methods:} Finally, we compare against state-of-the-art inference-time intervention methods including ITI~\citep{ref:li2024inference}, ICV~\citep{ref:liu2023context}, NL-ITI~\citep{ref:hoscilowicz2024non}, and LITO~\citep{ref:bayat2024enhanced}, which test the effectiveness of dynamically modifying internal activations as opposed to directly altering model weights.
\end{itemize}

We provide more details on the data train-dev-test split for each dataset, hyperparameters for baselines, and \method{} in~\cref{sec:app-setting}.

\subsection{Main Results}

\begin{figure*}[t]
    \centering
    \includegraphics[width=0.9\linewidth]{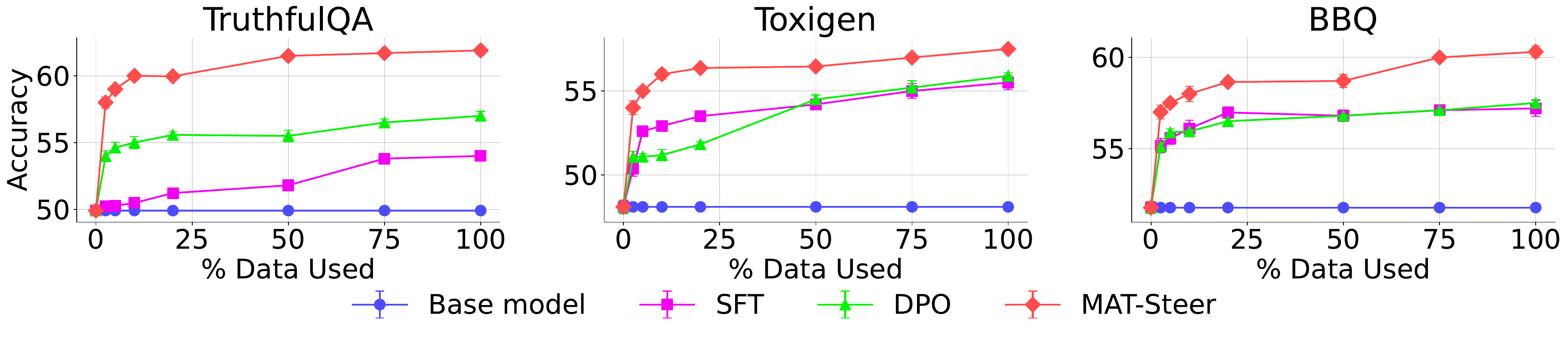}
    \caption{Accuracy on QA tasks versus the amount of training data on Llama-3.1-8B. Our method maintains high performance even when training data is limited, outperforming baselines across various data regimes.}
    \label{fig:data_scaling}
\end{figure*}

\begin{table}[t]
\small
\centering
\begin{tabular}{lccc}
\toprule
\textbf{Method}   & \textbf{TruthfulQA} & \textbf{Toxigen} & \textbf{BBQ} \\ \midrule
Llama-3.1-8B        & 49.91               & 48.10            & 51.77        \\ \midrule
ICL & 55.32               &   51.26          &   56.46      \\ \midrule
SFT              & 54.02               &   55.51          &     57.29    \\
DPO              & 56.10              &    55.94         &    57.51     \\ \midrule
Merge & 53.26    &   54.65          &    55.38     \\ 
RAdapt & 55.09               &   55.02          &    56.81     \\ 
\midrule
ITI              &   52.68               & 52.55            &   53.45    \\
ICV              & 55.21               & 53.61            &      54.86   \\
NL-ITI          & 56.67               & 54.73            &   56.46      \\
LITO             & 58.63               & 54.08            &    58.14     \\ \midrule
\method{} \textit{(Ours)}          & \textbf{61.94}      & \textbf{57.59}   & \textbf{60.32} \\ 
\bottomrule
\end{tabular}
\caption{Performance comparison of \method{} against in-context learning, fine-tuning, multiple adapters, and intervention methods on Llama-3.1-8B. Each method is evaluated on three datasets: TruthfulQA, Toxigen, and BBQ. The highest performance for each dataset is highlighted in bold.}
\label{tab:qa-tasks}
\end{table}

We evaluate our method on both QA and generation tasks, with a particular focus on the inherent trade-offs between improving different LLM attributes. Our results show that, while each baseline offers improvements in certain attributes, \method{} strikes a more favorable balance by delivering consistent gains across all evaluated metrics.

\paragraph{\method{} on QA Tasks.} Table~\ref{tab:qa-tasks} presents multiple-choice accuracy on the TruthfulQA, Toxigen, and BBQ datasets, which respectively assess truthfulness, toxicity, and bias. Notably, \method{} achieves the highest performance on all three datasets, with accuracies of 61.94\% (TruthfulQA), 57.59\% (Toxigen), and 60.32\% (BBQ). In contrast, fine-tuning approaches (e.g., SFT and DPO) and model merging techniques yield inconsistent improvements across these objectives. For instance, while fine-tuning with LoRA adapters may boost performance on one dataset, it fails to generalize across all targeted attributes. Among ITI methods, LITO demonstrates strong performance; however, 
\method{} still outperforms LITO by a significant margin, improving accuracy by 3.31\% on TruthfulQA, 3.51\% on Toxigen, and 2.18\% on BBQ. These results show that \method{} effectively balances different attributes and improves the trustworthiness of LLM outputs.

\paragraph{\method{} Generates Correct, Helpful, and Coherent Response.}  
For generation tasks, we evaluate our approach using the HelpSteer dataset~\citep{ref:wang2023helpsteer, ref:wang2024helpsteer2}, which is designed to align LLM outputs with human-preferred characteristics such as helpfulness, correctness, coherence, complexity, and verbosity.
Here, following~\citet{ref:zheng2023judging, ref:thakur2024judging}, each response is scored by GPT-4o on each attribute, and we compute win rates by comparing the overall average attribute scores. As shown in~\cref{fig:gen-tasks}, \method{} consistently achieves higher win rates compared to all baselines. 
This not only demonstrates that \method{} enhances the desired attributes (e.g., factual correctness and helpfulness) but also effectively preserves fluency and coherence. 

\paragraph{\method{} is more Data Efficient than LoRA Fine-Tuning.}
Results in~\cref{tab:qa-tasks} show that \method{} outperforms LoRA fine-tuning on the full dataset. 
To show \method{}'s effectiveness under limited data scenarios, we gradually reduce the amount of training data and measure the corresponding performance. \cref{fig:data_scaling} plots performance (e.g., accuracy for QA tasks) versus the amount of training data available. \method{} consistently outperforms fine-tuning baselines even with a reduced training set. In particular, \method{} with less than $20\%$ training data achieves the same or better performance than SFT and DPO on the full training set. For example, with $10\%$ of the data on TruthfulQA, \method{} achieves better performance than both DPO (60.05\% and 55.98\%) and SFT (60.05\% and 54.12\%) with $100\%$ of the data. 

\section{Analysis} \label{sec:main-analysis}

In this section, we provide a comprehensive analysis of our method, focusing on the internal mechanism, the trade-offs in intervention and demonstrating the robustness and generalization capabilities of our approach. \modified{Additionally, we provide an ablation study for different components of \method{}.}

\paragraph{Internal Mechanism of \method{}.} \modified{To evaluate how \method{} generalizes and adapts to both negative samples (where strong intervention is needed) and positive samples (where minimal or no intervention is needed) for a specific task, we conduct an analysis focused on toxicity. After obtaining steering vectors for~\cref{tab:qa-tasks} for Llama-3.1-8B, following the setup in ICV~\citep{ref:liu2023context}, we use the ParaDetox dataset~\citep{ref:logacheva2022paradetox} to examine how samples interact with the toxicity steering vector. We randomly sample 100 toxic samples (requiring intervention) and 100 neutral samples (no intervention expected). For each token in every sample, we record the gating weight for the toxicity vector as well as for other attribute vectors (e.g., truthfulness, bias) and the average number of intervened tokens. In addition, we follow \citet{ref:liu2023context}'s evaluation method by measuring the percentage of toxic samples that flip to neutral (higher is better) after applying MAT-Steer.}

\modified{\cref{tab:paradetox} shows that MAT-Steer correctly selects the toxicity vector with high gating weight, while keeping unrelated attributes largely inactive (0.61 vs. 0.14), demonstrating the importance of sparsity and orthogonality in ensuring targeted steering. This leads to 86\% toxicity decrease in ParaDetox. For neutral samples, the gating weight and the number of intervened tokens remain low across all attributes (0.08 and 0.12), showing \method{}'s ability to preserve aligned outputs without unnecessary intervention. 
}

\begin{table}[h]
\small
\centering
\begin{tabular}{lcc}
\toprule
\textbf{Metric} & \textbf{Toxic} & \textbf{Non-Toxic} \\
\midrule
Avg Gating (Toxicity) & \textbf{0.61} & 0.08 \\
Avg Gating (Other Attributes) & 0.14 & 0.12 \\
Toxicity Reduction & \textbf{-86\%} & - \\
Avg \# of Intervened Tokens & 3.9 & 0.6 \\
\bottomrule
\end{tabular}
\caption{Analysis of \method{} on ParaDetox.}
\label{tab:paradetox}
\end{table}

\begin{figure}[t]
    \centering
    \includegraphics[width=0.95\linewidth]{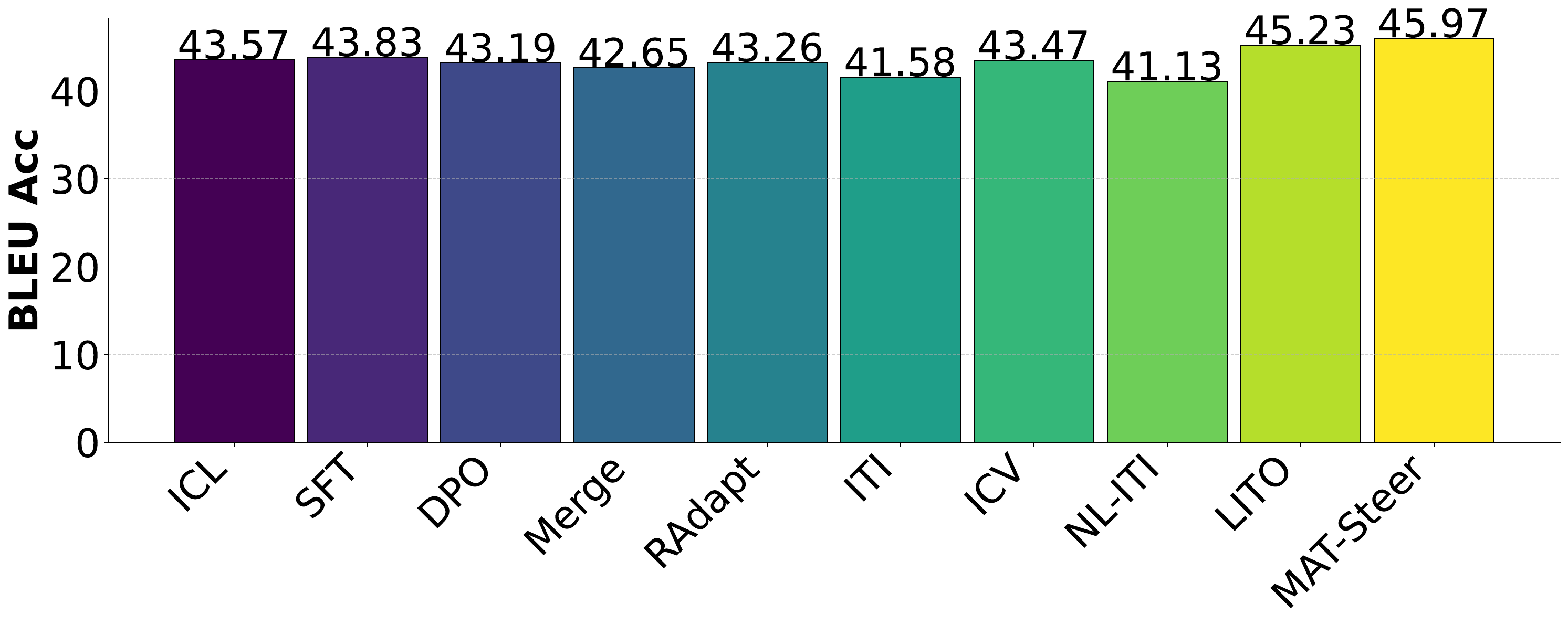}
    \caption{Comparing of BLEU score of \method{} with baselines on the generation split of TruthfulQA.
    }
    \label{fig:bleu}
\end{figure}

\paragraph{Impact on General LLM Capabilities.} To evaluate the impact of our intervention on text generation fluency, we follow a previous work in activation intervention ~\citep{ref:pham2024householder} and conduct open-ended generation experiments on TruthfulQA using Llama-3.1-8B. 
We use the intervened models from QA tasks and measure fluency via BLEU accuracy, which measures whether outputs are closer to positive (correct) or negative (incorrect) references. 
As shown in~\cref{fig:bleu}, \method{} yields higher BLEU accuracy than LoRA fine-tuning (e.g., 45.97 vs. 43.83 SFT) and ITI (45.97 vs. 41.58), indicating more factually correct and coherent outputs. 

\begin{figure}[t]
    \centering
    \includegraphics[width=0.95\linewidth]{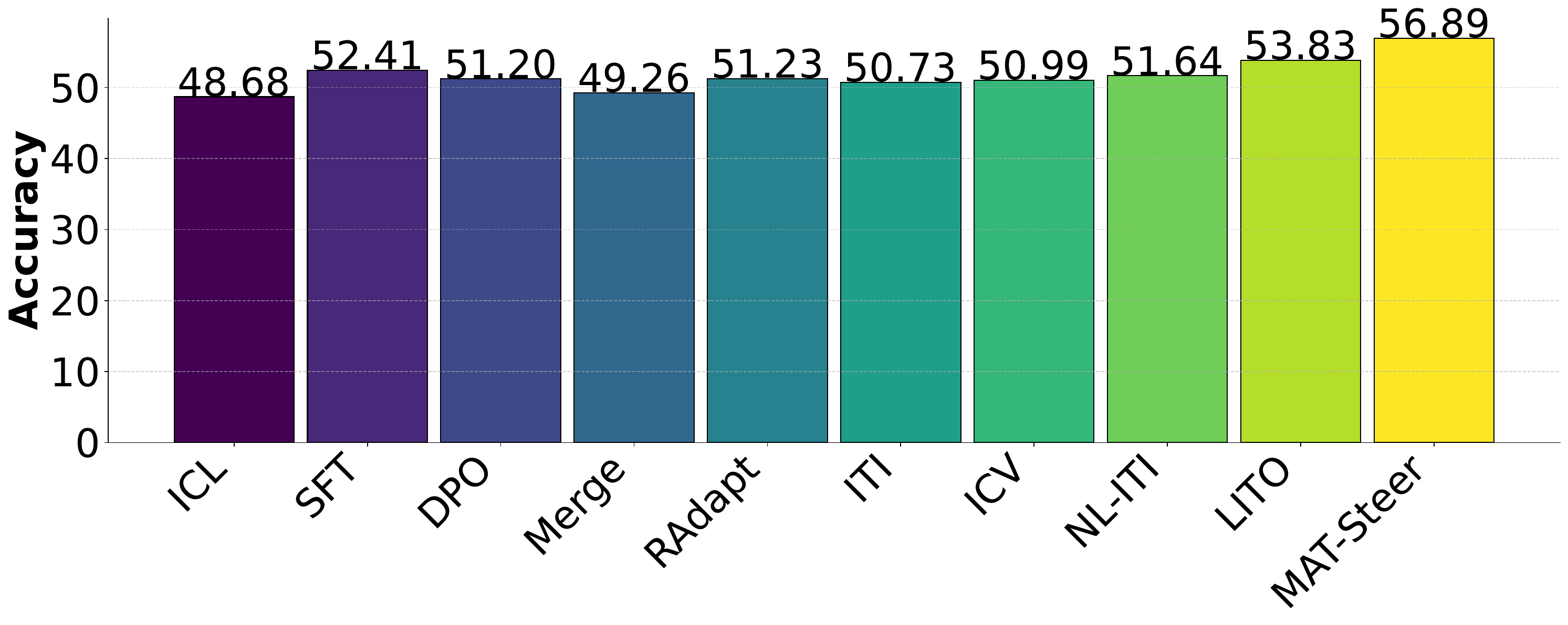}
    \caption{Generalization of different baselines and \method{} on FaithEval counterfactual dataset.}
    \label{fig:faitheval}
\end{figure}

\paragraph{Generalization to Other Tasks.} To further illustrate the generalization capability and interpretability advantages of our gating mechanism, we conduct experiments on the FaithEval~\citep{ref:ming2024faitheval} counterfactual dataset, a contextual QA benchmark designed to assess model faithfulness. This dataset presents questions with counterfactual context (statements that contradict common sense or widely accepted facts), challenging models to maintain robustness against misleading information. Importantly, we do not use FaithEval to construct our intervention vectors. 
Rather, we evaluate our pretrained and intervened models on it. As shown in Figure~\ref{fig:faitheval}, our method achieves the highest accuracy of 56.89\%, surpassing baselines such as ICL (48.68\%) and DPO (51.20\%). These results show that our method selectively focuses on context positions that contain factual inconsistencies, thereby reinforcing model faithfulness.

\paragraph{Ablation Study of \method{}.} \modified{Table~\ref{tab:ablation} presents an ablation study on the key components of \method{} evaluated on the TruthfulQA test set. Starting with the base model, which achieves 49.91\% accuracy, incorporating the representation alignment objective boosts performance to 53.82\%. Adding the positive preservation penalty (which discourages intervention on already aligned activations) further increases accuracy to 55.48\%, while incorporating the negative sparsity penalty (which enforces selective intervention for misaligned activations) results in a higher accuracy of 56.73\%. Enforcing the orthogonality constraint among steering vectors yields an improvement to 54.37\%. Moreover, we observe that normalization plays a crucial role: an ablated version of \method{} without normalization achieves 59.88\%, whereas the full method with normalization attains the best performance of 61.94\% accuracy. Finally, an ablated version of \method{} without any components in~\cref{sec:method} leads to a significant drop in performance (e.g., 3.86\% drop without positive preservation). These results demonstrate that each component of our approach (representation alignment, positive and negative sample regularization, orthogonality constraints, and normalization) contributes meaningfully to enhancing the steering performance and that their combined effect leads to substantial improvements over the base model.}

\begin{table}[h]
\small
\centering
\setlength{\tabcolsep}{0.5 pt}
\begin{tabular}{lccc}
\toprule
\textbf{Method}   & \textbf{TruthfulQA}  \\ \midrule
Llama-3.1-8B       & 49.91                     \\ 
\midrule
Llama-3.1-8B + Alignment & 53.82 \\
Llama-3.1-8B + Alignment + Pos &       55.48              \\ 
Llama-3.1-8B + Alignment + Sparse              &      56.73               \\
Llama-3.1-8B + Alignment + Orth          &      54.37               \\
\midrule
\method{} w/o Pos          &      57.37              \\
\method{} w/o Sparse          &      58.08              \\
\method{} w/o Orth          &      59.69             \\
\method{} w/o Normalization          &      59.88              \\
\midrule
\method{}             & \textbf{61.94}       \\ 
\bottomrule
\end{tabular}
\caption{Ablation study on different components of \method{} on Llama-3.1-8B.}
\label{tab:ablation}
\end{table}

\paragraph{Additional Results and Analyses.} In the appendix, we provide extensive additional analyses of \method{} along with more numerical results. 
In particular, in~\cref{fig:scaling}, we demonstrate how the performance of \method{} scales as the number of attributes increases, with \emph{increasing} win rates over baselines as more attributes are added.
Furthermore, we compare various token intervention methods in~\cref{tab:token-seletion}, finding that \method{}'s improvements are due to selecting the \emph{right} tokens (not just fewer tokens). 
In~\cref{sec:app-results}, we present further results, including an additional safety benchmark (HH-RLHF) and a reasoning dataset (OBQA) (see~\cref{tab:hh-rlhf} and~\cref{tab:obqa}), and generalization across different model families (see~\cref{tab:qa-tasks-chat} and~\cref{tab:qa-tasks-qwen}) -- where \method{} also outperforms all baselines -- 
as well as results showing that our method can be effectively combined with ICL and fine-tuning approaches (see~\cref{tab:fsp-iti}), further enhancing these methods. 
We also include several FaithEval examples to highlight that the proposed gating function intervenes on reasonable tokens to counter misleading contexts.

\section{Related Work}
Recent advancements in LLMs for multi-task and multi-attribute applications have explored several techniques, including prompting and reinforcement learning from human feedback (RLHF). 

\paragraph{Multi-task Prompting.} Prompting-based techniques design specialized inputs to guide LLMs toward desired attributes across a range of tasks~\citep{ref:li2021prefix, ref:qin2021learning, ref:prasad2023grips}. For example, prompt tuning methods aim to learn a shared prompt that adapts to multiple tasks~\citep{ref:liu2023hierarchical, ref:tian2024argue, ref:kim2024aapl}, while \citet{ref:xu2025mixture} employs instruction concatenation combined with diverse system prompts to improve alignment efficiency in applications such as dialogue and coding, and mathematical reasoning.

\paragraph{Multi-task RLHF.} RLHF aims to train LLMs to align their outputs with human preferences~\citep{ref:ouyang2022training, ref:rafailov2024direct}. Although RLHF has shown promise in adjusting model behavior, previous work by~\citet{ref:dong2023abilities, ref:kotha2024understanding, ref:biderman2024lora} show that multi-task fine-tuning can lead to conflicts between different objectives. To mitigate such issues, several lines of work have proposed methods to balance these competing goals. For instance, \citet{ref:liu2024mftcoder} address data bias among multiple coding tasks, and other efforts focus on preference fine-tuning with multiple objectives for improving LLM helpfulness~\citep{ref:wu2023finegrained, ref:zhang2024bi, ref:yang2024metaaligner, ref:wang2024arithmetic, ref:wang2024interpretable, nguyen2025laser}.

In contrast to these approaches, which primarily tune either the input prompt or the model parameters, our work focuses on inference-time intervention in the representation space of LLMs under a multi-attribute setting. 
Moreover, our experiments show that applying our intervention on top of prompting or fine-tuning methods further enhances model performance (see~\cref{sec:app-results}).

\section{Conclusion}
We introduced \method{}, a novel and parameter-efficient approach for inference-time intervention that dynamically steers large language models according to multiple potentially conflicting attributes. 
By leveraging a gating mechanism and a new optimization objective, \method{} selectively adjusts representations at the token level to mitigate undesirable outputs while preserving overall model capabilities. Extensive experiments demonstrate that \method{} outperforms existing approaches across a range of tasks, achieving improved accuracy, better alignment, and robust generalization with significantly less training data.

\section*{Limitations}
Our method -- like the other intervention and trained baselines we compare it to -- may struggle in scenarios where the attributes to be aligned are highly complex or unsteerable~\citep{ref:tan2024analysing}. Moreover, like other steering methods, \method{} relies on a small set of data to create steering vectors; we show that \method{} is more data-efficient than several baseline methods, mitigating this issue. In addition, our evaluations have so far been conducted on a select set of tasks and attributes chosen to be representative of standard steering objectives. We leave expanding our approach to an even wider array of tasks as a direction for future research. Our work aims to mitigate LLM risks like bias and toxicity while maintaining performance on other behaviors like helpfulness. As such, \method{} mitigates some of the risks associated with LLMs; however, like other steering methods, \method{} does not eliminate these risks entirely.

\section*{Acknowledgments}
This work was supported by NSF-CAREER Award 1846185, the Microsoft Accelerate Foundation Models Research (AFMR) grant program, DARPA ECOLE Program No. HR00112390060, and NSF-AI Engage Institute DRL-2112635. 
Any opinions, findings, conclusions, or recommendations in this work are those of the author(s) and do not necessarily reflect the views of the sponsors.

\bibliography{bibliography}

@article{ref:zou2023representation,
  title={Representation engineering: A top-down approach to ai transparency},
  author={Zou, Andy and Phan, Long and Chen, Sarah and Campbell, James and Guo, Phillip and Ren, Richard and Pan, Alexander and Yin, Xuwang and Mazeika, Mantas and Dombrowski, Ann-Kathrin and others},
  journal={arXiv preprint arXiv:2310.01405},
  year={2023}
}

@article{ref:li2024inference,
  title={Inference-time intervention: Eliciting truthful answers from a language model},
  author={Li, Kenneth and Patel, Oam and Vi{\'e}gas, Fernanda and Pfister, Hanspeter and Wattenberg, Martin},
  journal={Advances in Neural Information Processing Systems},
  volume={36},
  year={2024}
}

@inproceedings{ref:liu2023context,
  title={In-context Vectors: Making In Context Learning More Effective and Controllable Through Latent Space Steering},
  author={Liu, Sheng and Ye, Haotian and Xing, Lei and Zou, James Y},
  booktitle={Forty-first International Conference on Machine Learning},
  year={2024}
}

@article{ref:rafailov2024direct,
  title={Direct preference optimization: Your language model is secretly a reward model},
  author={Rafailov, Rafael and Sharma, Archit and Mitchell, Eric and Manning, Christopher D and Ermon, Stefano and Finn, Chelsea},
  journal={Advances in Neural Information Processing Systems},
  volume={36},
  year={2024}
}

@article{ref:van2024extending,
  title={Extending Activation Steering to Broad Skills and Multiple Behaviours},
  author={van der Weij, Teun and Poesio, Massimo and Schoots, Nandi},
  journal={arXiv preprint arXiv:2403.05767},
  year={2024}
}

@inproceedings{ref:ming2024faitheval,
    title={FaithEval: Can Your Language Model Stay Faithful to Context, Even If ''The Moon is Made of Marshmallows''},
    author={Yifei Ming and Senthil Purushwalkam and Shrey Pandit and Zixuan Ke and Xuan-Phi Nguyen and Caiming Xiong and Shafiq Joty},
    booktitle={The Thirteenth International Conference on Learning Representations},
    year={2025},
    url={https://openreview.net/forum?id=UeVx6L59fg}
}

@inproceedings{ref:wu2024reft,
title={Re{FT}: Representation Finetuning for Language Models},
author={Zhengxuan Wu and Aryaman Arora and Zheng Wang and Atticus Geiger and Dan Jurafsky and Christopher D Manning and Christopher Potts},
booktitle={The Thirty-eighth Annual Conference on Neural Information Processing Systems},
year={2024},
url={https://openreview.net/forum?id=fykjplMc0V}
}

@inproceedings{ref:wang2024interpretable,
    title = "Interpretable Preferences via Multi-Objective Reward Modeling and Mixture-of-Experts",
    author = "Wang, Haoxiang  and
      Xiong, Wei  and
      Xie, Tengyang  and
      Zhao, Han  and
      Zhang, Tong",
    editor = "Al-Onaizan, Yaser  and
      Bansal, Mohit  and
      Chen, Yun-Nung",
    booktitle = "Findings of the Association for Computational Linguistics: EMNLP 2024",
    month = nov,
    year = "2024",
    address = "Miami, Florida, USA",
    publisher = "Association for Computational Linguistics",
    url = "https://aclanthology.org/2024.findings-emnlp.620/",
    doi = "10.18653/v1/2024.findings-emnlp.620",
    pages = "10582--10592",
}

@inproceedings{ref:wang2024arithmetic,
    title = "Arithmetic Control of {LLM}s for Diverse User Preferences: Directional Preference Alignment with Multi-Objective Rewards",
    author = "Wang, Haoxiang  and
      Lin, Yong  and
      Xiong, Wei  and
      Yang, Rui  and
      Diao, Shizhe  and
      Qiu, Shuang  and
      Zhao, Han  and
      Zhang, Tong",
    editor = "Ku, Lun-Wei  and
      Martins, Andre  and
      Srikumar, Vivek",
    booktitle = "Proceedings of the 62nd Annual Meeting of the Association for Computational Linguistics (Volume 1: Long Papers)",
    month = aug,
    year = "2024",
    address = "Bangkok, Thailand",
    publisher = "Association for Computational Linguistics",
    url = "https://aclanthology.org/2024.acl-long.468/",
    doi = "10.18653/v1/2024.acl-long.468",
    pages = "8642--8655",
}

@article{ref:ouyang2022training,
  title={Training language models to follow instructions with human feedback},
  author={Ouyang, Long and Wu, Jeffrey and Jiang, Xu and Almeida, Diogo and Wainwright, Carroll and Mishkin, Pamela and Zhang, Chong and Agarwal, Sandhini and Slama, Katarina and Ray, Alex and others},
  journal={Advances in neural information processing systems},
  volume={35},
  pages={27730--27744},
  year={2022}
}

@inproceedings{ref:hu2021lora,
    title={Lo{RA}: Low-Rank Adaptation of Large Language Models},
    author={Edward J Hu and yelong shen and Phillip Wallis and Zeyuan Allen-Zhu and Yuanzhi Li and Shean Wang and Lu Wang and Weizhu Chen},
    booktitle={International Conference on Learning Representations},
    year={2022},
    url={https://openreview.net/forum?id=nZeVKeeFYf9}
}

@misc{ref:qwen2024moe,
    title = {Qwen2.5: A Party of Foundation Models},
    url = {https://qwenlm.github.io/blog/qwen2.5/},
    author = {Qwen Team},
    month = {September},
    year = {2024}
}

@article{ref:achiam2023gpt,
  title={Gpt-4 technical report},
  author={Achiam, Josh and Adler, Steven and Agarwal, Sandhini and Ahmad, Lama and Akkaya, Ilge and Aleman, Florencia Leoni and Almeida, Diogo and Altenschmidt, Janko and Altman, Sam and Anadkat, Shyamal and others},
  journal={arXiv preprint arXiv:2303.08774},
  year={2023}
}

@article{ref:dubey2024llama,
  title={The llama 3 herd of models},
  author={Dubey, Abhimanyu and Jauhri, Abhinav and Pandey, Abhinav and Kadian, Abhishek and Al-Dahle, Ahmad and Letman, Aiesha and Mathur, Akhil and Schelten, Alan and Yang, Amy and Fan, Angela and others},
  journal={arXiv preprint arXiv:2407.21783},
  year={2024}
}

@inproceedings{ref:huang2024trustllm,
  title={TrustLLM: Trustworthiness in Large Language Models},
  author={Yue Huang and Lichao Sun and Haoran Wang and Siyuan Wu and Qihui Zhang and Yuan Li and Chujie Gao and Yixin Huang and Wenhan Lyu and Yixuan Zhang and Xiner Li and Hanchi Sun and Zhengliang Liu and Yixin Liu and Yijue Wang and Zhikun Zhang and Bertie Vidgen and Bhavya Kailkhura and Caiming Xiong and Chaowei Xiao and Chunyuan Li and Eric P. Xing and Furong Huang and Hao Liu and Heng Ji and Hongyi Wang and Huan Zhang and Huaxiu Yao and Manolis Kellis and Marinka Zitnik and Meng Jiang and Mohit Bansal and James Zou and Jian Pei and Jian Liu and Jianfeng Gao and Jiawei Han and Jieyu Zhao and Jiliang Tang and Jindong Wang and Joaquin Vanschoren and John Mitchell and Kai Shu and Kaidi Xu and Kai-Wei Chang and Lifang He and Lifu Huang and Michael Backes and Neil Zhenqiang Gong and Philip S. Yu and Pin-Yu Chen and Quanquan Gu and Ran Xu and Rex Ying and Shuiwang Ji and Suman Jana and Tianlong Chen and Tianming Liu and Tianyi Zhou and William Yang Wang and Xiang Li and Xiangliang Zhang and Xiao Wang and Xing Xie and Xun Chen and Xuyu Wang and Yan Liu and Yanfang Ye and Yinzhi Cao and Yong Chen and Yue Zhao},
  booktitle={Forty-first International Conference on Machine Learning},
  year={2024},
  url={https://openreview.net/forum?id=bWUU0LwwMp}
}

@inproceedings{ref:sidharth2024controlled,
  author={Sidharth Mudgal and Jong Lee and Harish Ganapathy and YaGuang Li and Tao Wang and Yanping Huang and Zhifeng Chen and Heng-Tze Cheng and Michael Collins and Trevor Strohman and Jilin Chen and Alex Beutel and Ahmad Beirami},
  title={Controlled Decoding from Language Models},
  year={2024},
  cdate={1704067200000},
  url={https://openreview.net/forum?id=bVIcZb7Qa0},
  booktitle={ICML},
}

@inproceedings{ref:shi2024decoding,
    title={Decoding-Time Language Model Alignment with Multiple Objectives},
    author={Ruizhe Shi and Yifang Chen and Yushi Hu and Alisa Liu and Hannaneh Hajishirzi and Noah A. Smith and Simon Shaolei Du},
    booktitle={ICML 2024 Workshop on Theoretical Foundations of Foundation Models},
    year={2024},
    url={https://openreview.net/forum?id=RmGvEmttB7}
}

@article{ref:rame2024rewarded,
  title={Rewarded soups: towards pareto-optimal alignment by interpolating weights fine-tuned on diverse rewards},
  author={Rame, Alexandre and Couairon, Guillaume and Dancette, Corentin and Gaya, Jean-Baptiste and Shukor, Mustafa and Soulier, Laure and Cord, Matthieu},
  journal={Advances in Neural Information Processing Systems},
  volume={36},
  year={2024}
}

@inproceedings{ref:shaikh2022second,
    title = "On Second Thought, Let`s Not Think Step by Step! Bias and Toxicity in Zero-Shot Reasoning",
    author = "Shaikh, Omar  and
      Zhang, Hongxin  and
      Held, William  and
      Bernstein, Michael  and
      Yang, Diyi",
    editor = "Rogers, Anna  and
      Boyd-Graber, Jordan  and
      Okazaki, Naoaki",
    booktitle = "Proceedings of the 61st Annual Meeting of the Association for Computational Linguistics (Volume 1: Long Papers)",
    month = jul,
    year = "2023",
    address = "Toronto, Canada",
    publisher = "Association for Computational Linguistics",
    url = "https://aclanthology.org/2023.acl-long.244/",
    doi = "10.18653/v1/2023.acl-long.244",
    pages = "4454--4470",
}

@inproceedings{ref:rimsky2024steering,
    title = "Steering Llama 2 via Contrastive Activation Addition",
    author = "Rimsky, Nina  and
      Gabrieli, Nick  and
      Schulz, Julian  and
      Tong, Meg  and
      Hubinger, Evan  and
      Turner, Alexander",
    editor = "Ku, Lun-Wei  and
      Martins, Andre  and
      Srikumar, Vivek",
    booktitle = "Proceedings of the 62nd Annual Meeting of the Association for Computational Linguistics (Volume 1: Long Papers)",
    month = aug,
    year = "2024",
    address = "Bangkok, Thailand",
    publisher = "Association for Computational Linguistics",
    url = "https://aclanthology.org/2024.acl-long.828/",
    doi = "10.18653/v1/2024.acl-long.828",
    pages = "15504--15522",
}

@misc{ref:turner2024steering,
      title={Steering Language Models With Activation Engineering}, 
      author={Alexander Matt Turner and Lisa Thiergart and Gavin Leech and David Udell and Juan J. Vazquez and Ulisse Mini and Monte MacDiarmid},
      year={2024},
      eprint={2308.10248},
      archivePrefix={arXiv},
      primaryClass={cs.CL},
      url={https://arxiv.org/abs/2308.10248}, 
}

@inproceedings{ref:hollinsworth2024language,
    title = "Language Models Linearly Represent Sentiment",
    author = "Hollinsworth, Oskar John  and
      Tigges, Curt  and
      Geiger, Atticus  and
      Nanda, Neel",
    editor = "Belinkov, Yonatan  and
      Kim, Najoung  and
      Jumelet, Jaap  and
      Mohebbi, Hosein  and
      Mueller, Aaron  and
      Chen, Hanjie",
    booktitle = "Proceedings of the 7th BlackboxNLP Workshop: Analyzing and Interpreting Neural Networks for NLP",
    month = nov,
    year = "2024",
    address = "Miami, Florida, US",
    publisher = "Association for Computational Linguistics",
    url = "https://aclanthology.org/2024.blackboxnlp-1.5/",
    doi = "10.18653/v1/2024.blackboxnlp-1.5",
    pages = "58--87",
}

@article{ref:biderman2024lora,
title={Lo{RA} Learns Less and Forgets Less},
author={Dan Biderman and Jacob Portes and Jose Javier Gonzalez Ortiz and Mansheej Paul and Philip Greengard and Connor Jennings and Daniel King and Sam Havens and Vitaliy Chiley and Jonathan Frankle and Cody Blakeney and John Patrick Cunningham},
journal={Transactions on Machine Learning Research},
issn={2835-8856},
year={2024},
url={https://openreview.net/forum?id=aloEru2qCG},
note={Featured Certification}
}

@inproceedings{ref:kotha2024understanding,
title={Understanding Catastrophic Forgetting in Language Models via Implicit Inference},
author={Suhas Kotha and Jacob Mitchell Springer and Aditi Raghunathan},
booktitle={The Twelfth International Conference on Learning Representations},
year={2024},
url={https://openreview.net/forum?id=VrHiF2hsrm}
}

@inproceedings{ref:tan2024analysing,
title={Analysing the Generalisation and Reliability of Steering Vectors},
author={Daniel Chee Hian Tan and David Chanin and Aengus Lynch and Brooks Paige and Dimitrios Kanoulas and Adri{\`a} Garriga-Alonso and Robert Kirk},
booktitle={The Thirty-eighth Annual Conference on Neural Information Processing Systems},
year={2024},
url={https://openreview.net/forum?id=v8X70gTodR}
}

@inproceedings{ref:lin2021truthfulqa,
    title = "{T}ruthful{QA}: Measuring How Models Mimic Human Falsehoods",
    author = "Lin, Stephanie  and
      Hilton, Jacob  and
      Evans, Owain",
    editor = "Muresan, Smaranda  and
      Nakov, Preslav  and
      Villavicencio, Aline",
    booktitle = "Proceedings of the 60th Annual Meeting of the Association for Computational Linguistics (Volume 1: Long Papers)",
    month = may,
    year = "2022",
    address = "Dublin, Ireland",
    publisher = "Association for Computational Linguistics",
    url = "https://aclanthology.org/2022.acl-long.229/",
    doi = "10.18653/v1/2022.acl-long.229",
    pages = "3214--3252",
}

@inproceedings{ref:hartvigsen2022toxigen,
    title = "{T}oxi{G}en: A Large-Scale Machine-Generated Dataset for Adversarial and Implicit Hate Speech Detection",
    author = "Hartvigsen, Thomas  and
      Gabriel, Saadia  and
      Palangi, Hamid  and
      Sap, Maarten  and
      Ray, Dipankar  and
      Kamar, Ece",
    editor = "Muresan, Smaranda  and
      Nakov, Preslav  and
      Villavicencio, Aline",
    booktitle = "Proceedings of the 60th Annual Meeting of the Association for Computational Linguistics (Volume 1: Long Papers)",
    month = may,
    year = "2022",
    address = "Dublin, Ireland",
    publisher = "Association for Computational Linguistics",
    url = "https://aclanthology.org/2022.acl-long.234/",
    doi = "10.18653/v1/2022.acl-long.234",
    pages = "3309--3326",
}

@inproceedings{ref:wang2023helpsteer,
    title = "{H}elp{S}teer: Multi-attribute Helpfulness Dataset for {S}teer{LM}",
    author = "Wang, Zhilin  and
      Dong, Yi  and
      Zeng, Jiaqi  and
      Adams, Virginia  and
      Sreedhar, Makesh Narsimhan  and
      Egert, Daniel  and
      Delalleau, Olivier  and
      Scowcroft, Jane  and
      Kant, Neel  and
      Swope, Aidan  and
      Kuchaiev, Oleksii",
    editor = "Duh, Kevin  and
      Gomez, Helena  and
      Bethard, Steven",
    booktitle = "Proceedings of the 2024 Conference of the North American Chapter of the Association for Computational Linguistics: Human Language Technologies (Volume 1: Long Papers)",
    month = jun,
    year = "2024",
    address = "Mexico City, Mexico",
    publisher = "Association for Computational Linguistics",
    url = "https://aclanthology.org/2024.naacl-long.185/",
    doi = "10.18653/v1/2024.naacl-long.185",
    pages = "3371--3384",
}

@inproceedings{ref:dong2023steerlm,
    title={Steer{LM}: Attribute Conditioned {SFT} as an (User-Steerable) Alternative to {RLHF}},
    author={Yi Dong and Zhilin Wang and Makesh Narsimhan Sreedhar and Xianchao Wu and Oleksii Kuchaiev},
    booktitle={The 2023 Conference on Empirical Methods in Natural Language Processing},
    year={2023},
    url={https://openreview.net/forum?id=J5FFUHZjNx}
}

@article{ref:brown2020language,
  title={Language models are few-shot learners},
  author={Brown, Tom and Mann, Benjamin and Ryder, Nick and Subbiah, Melanie and Kaplan, Jared D and Dhariwal, Prafulla and Neelakantan, Arvind and Shyam, Pranav and Sastry, Girish and Askell, Amanda and others},
  journal={Advances in neural information processing systems},
  volume={33},
  pages={1877--1901},
  year={2020}
}

@inproceedings{ref:hoscilowicz2024non,
  title={Non-Linear Inference Time Intervention: Improving LLM Truthfulness},
  author={Hoscilowicz, Jakub and Wiacek, Adam and Chojnacki, Jan and Cieslak, Adam and Michon, Leszek and Janicki, Artur},
  booktitle={Proc. Interspeech 2024},
  pages={4094--4098},
  year={2024}
}

@article{ref:gretton2012kernel,
  title={A kernel two-sample test},
  author={Gretton, Arthur and Borgwardt, Karsten M and Rasch, Malte J and Sch{\"o}lkopf, Bernhard and Smola, Alexander},
  journal={The Journal of Machine Learning Research},
  volume={13},
  number={1},
  pages={723--773},
  year={2012},
  publisher={JMLR. org}
}

@inproceedings{ref:wang2024helpsteer2,
title={HelpSteer 2: Open-source dataset for training top-performing reward models},
author={Zhilin Wang and Yi Dong and Olivier Delalleau and Jiaqi Zeng and Gerald Shen and Daniel Egert and Jimmy J. Zhang and Makesh Narsimhan Sreedhar and Oleksii Kuchaiev},
booktitle={The Thirty-eight Conference on Neural Information Processing Systems Datasets and Benchmarks Track},
year={2024},
url={https://openreview.net/forum?id=PvVKUFhaNy}
}

@article{ref:zheng2023judging,
  title={Judging llm-as-a-judge with mt-bench and chatbot arena},
  author={Zheng, Lianmin and Chiang, Wei-Lin and Sheng, Ying and Zhuang, Siyuan and Wu, Zhanghao and Zhuang, Yonghao and Lin, Zi and Li, Zhuohan and Li, Dacheng and Xing, Eric and others},
  journal={Advances in Neural Information Processing Systems},
  volume={36},
  pages={46595--46623},
  year={2023}
}

@article{ref:thakur2024judging,
  title={Judging the Judges: Evaluating Alignment and Vulnerabilities in LLMs-as-Judges},
  author={Thakur, Aman Singh and Choudhary, Kartik and Ramayapally, Venkat Srinik and Vaidyanathan, Sankaran and Hupkes, Dieuwke},
  journal={arXiv preprint arXiv:2406.12624},
  year={2024}
}

@article{ref:li2017learning,
  title={Learning without forgetting},
  author={Li, Zhizhong and Hoiem, Derek},
  journal={IEEE transactions on pattern analysis and machine intelligence},
  volume={40},
  number={12},
  pages={2935--2947},
  year={2017},
  publisher={IEEE}
}

@article{ref:lopez2017gradient,
  title={Gradient episodic memory for continual learning},
  author={Lopez-Paz, David and Ranzato, Marc'Aurelio},
  journal={Advances in neural information processing systems},
  volume={30},
  year={2017}
}

@inproceedings{ref:pham2024householder,
    title = "Householder Pseudo-Rotation: A Novel Approach to Activation Editing in {LLM}s with Direction-Magnitude Perspective",
    author = "Pham, Van-Cuong  and
      Nguyen, Thien Huu",
    editor = "Al-Onaizan, Yaser  and
      Bansal, Mohit  and
      Chen, Yun-Nung",
    booktitle = "Proceedings of the 2024 Conference on Empirical Methods in Natural Language Processing",
    month = nov,
    year = "2024",
    address = "Miami, Florida, USA",
    publisher = "Association for Computational Linguistics",
    url = "https://aclanthology.org/2024.emnlp-main.761/",
    doi = "10.18653/v1/2024.emnlp-main.761",
    pages = "13737--13751"
}

@inproceedings{ref:liu2023hierarchical,
  title={Hierarchical prompt learning for multi-task learning},
  author={Liu, Yajing and Lu, Yuning and Liu, Hao and An, Yaozu and Xu, Zhuoran and Yao, Zhuokun and Zhang, Baofeng and Xiong, Zhiwei and Gui, Chenguang},
  booktitle={Proceedings of the IEEE/CVF Conference on Computer Vision and Pattern Recognition},
  pages={10888--10898},
  year={2023}
}

@inproceedings{ref:tian2024argue,
  title={ArGue: Attribute-Guided Prompt Tuning for Vision-Language Models},
  author={Tian, Xinyu and Zou, Shu and Yang, Zhaoyuan and Zhang, Jing},
  booktitle={Proceedings of the IEEE/CVF Conference on Computer Vision and Pattern Recognition},
  pages={28578--28587},
  year={2024}
}

@inproceedings{ref:kim2024aapl,
  title={AAPL: Adding Attributes to Prompt Learning for Vision-Language Models},
  author={Kim, Gahyeon and Kim, Sohee and Lee, Seokju},
  booktitle={Proceedings of the IEEE/CVF Conference on Computer Vision and Pattern Recognition},
  pages={1572--1582},
  year={2024}
}

@misc{ref:xu2025mixture,
      title={Mixture-of-Instructions: Aligning Large Language Models via Mixture Prompting}, 
      author={Bowen Xu and Shaoyu Wu and Kai Liu and Lulu Hu},
      year={2025},
      eprint={2404.18410},
      archivePrefix={arXiv},
      primaryClass={cs.CL},
      url={https://arxiv.org/abs/2404.18410}, 
}

@inproceedings{ref:dong2023abilities,
    title = "How Abilities in Large Language Models are Affected by Supervised Fine-tuning Data Composition",
    author = "Dong, Guanting  and
      Yuan, Hongyi  and
      Lu, Keming  and
      Li, Chengpeng  and
      Xue, Mingfeng  and
      Liu, Dayiheng  and
      Wang, Wei  and
      Yuan, Zheng  and
      Zhou, Chang  and
      Zhou, Jingren",
    editor = "Ku, Lun-Wei  and
      Martins, Andre  and
      Srikumar, Vivek",
    booktitle = "Proceedings of the 62nd Annual Meeting of the Association for Computational Linguistics (Volume 1: Long Papers)",
    month = aug,
    year = "2024",
    address = "Bangkok, Thailand",
    publisher = "Association for Computational Linguistics",
    url = "https://aclanthology.org/2024.acl-long.12/",
    doi = "10.18653/v1/2024.acl-long.12",
    pages = "177--198",
}

@inproceedings{ref:liu2024mftcoder,
  title={Mftcoder: Boosting code llms with multitask fine-tuning},
  author={Liu, Bingchang and Chen, Chaoyu and Gong, Zi and Liao, Cong and Wang, Huan and Lei, Zhichao and Liang, Ming and Chen, Dajun and Shen, Min and Zhou, Hailian and others},
  booktitle={Proceedings of the 30th ACM SIGKDD Conference on Knowledge Discovery and Data Mining},
  pages={5430--5441},
  year={2024}
}

@inproceedings{ref:yang2024metaaligner,
  title={Metaaligner: Towards generalizable multi-objective alignment of language models},
  author={Yang, Kailai and Liu, Zhiwei and Xie, Qianqian and Huang, Jimin and Zhang, Tianlin and Ananiadou, Sophia},
  booktitle={The Thirty-eighth Annual Conference on Neural Information Processing Systems},
  year={2024}
}

@inproceedings{ref:zhang2024bi,
    title={Bi-Factorial Preference Optimization: Balancing Safety-Helpfulness in Language Models},
    author={Wenxuan Zhang and Philip Torr and Mohamed Elhoseiny and Adel Bibi},
    booktitle={The Thirteenth International Conference on Learning Representations},
    year={2025},
    url={https://openreview.net/forum?id=GjM61KRiTG}
}

@inproceedings{ref:wu2023finegrained,
title={Fine-Grained Human Feedback Gives Better Rewards for Language Model Training},
author={Zeqiu Wu and Yushi Hu and Weijia Shi and Nouha Dziri and Alane Suhr and Prithviraj Ammanabrolu and Noah A. Smith and Mari Ostendorf and Hannaneh Hajishirzi},
booktitle={Thirty-seventh Conference on Neural Information Processing Systems},
year={2023},
url={https://openreview.net/forum?id=CSbGXyCswu}
}

@article{ref:yang2024moral,
  title={MoRAL: MoE Augmented LoRA for LLMs' Lifelong Learning},
  author={Yang, Shu and Ali, Muhammad Asif and Wang, Cheng-Long and Hu, Lijie and Wang, Di},
  journal={arXiv preprint arXiv:2402.11260},
  year={2024}
}

@inproceedings{ref:wortsman2022model,
  title={Model soups: averaging weights of multiple fine-tuned models improves accuracy without increasing inference time},
  author={Wortsman, Mitchell and Ilharco, Gabriel and Gadre, Samir Ya and Roelofs, Rebecca and Gontijo-Lopes, Raphael and Morcos, Ari S and Namkoong, Hongseok and Farhadi, Ali and Carmon, Yair and Kornblith, Simon and others},
  booktitle={International conference on machine learning},
  pages={23965--23998},
  year={2022},
  organization={PMLR}
}

@inproceedings{ref:prasad2023grips,
    title = "{G}r{IPS}: Gradient-free, Edit-based Instruction Search for Prompting Large Language Models",
    author = "Prasad, Archiki  and
      Hase, Peter  and
      Zhou, Xiang  and
      Bansal, Mohit",
    editor = "Vlachos, Andreas  and
      Augenstein, Isabelle",
    booktitle = "Proceedings of the 17th Conference of the European Chapter of the Association for Computational Linguistics",
    month = may,
    year = "2023",
    address = "Dubrovnik, Croatia",
    publisher = "Association for Computational Linguistics",
    url = "https://aclanthology.org/2023.eacl-main.277/",
    doi = "10.18653/v1/2023.eacl-main.277",
    pages = "3845--3864",
}

@inproceedings{ref:li2021prefix,
    title = "Prefix-Tuning: Optimizing Continuous Prompts for Generation",
    author = "Li, Xiang Lisa  and
      Liang, Percy",
    editor = "Zong, Chengqing  and
      Xia, Fei  and
      Li, Wenjie  and
      Navigli, Roberto",
    booktitle = "Proceedings of the 59th Annual Meeting of the Association for Computational Linguistics and the 11th International Joint Conference on Natural Language Processing (Volume 1: Long Papers)",
    month = aug,
    year = "2021",
    address = "Online",
    publisher = "Association for Computational Linguistics",
    url = "https://aclanthology.org/2021.acl-long.353/",
    doi = "10.18653/v1/2021.acl-long.353",
    pages = "4582--4597",
}

@inproceedings{ref:qin2021learning,
    title = "Learning How to Ask: Querying {LM}s with Mixtures of Soft Prompts",
    author = "Qin, Guanghui  and
      Eisner, Jason",
    editor = "Toutanova, Kristina  and
      Rumshisky, Anna  and
      Zettlemoyer, Luke  and
      Hakkani-Tur, Dilek  and
      Beltagy, Iz  and
      Bethard, Steven  and
      Cotterell, Ryan  and
      Chakraborty, Tanmoy  and
      Zhou, Yichao",
    booktitle = "Proceedings of the 2021 Conference of the North American Chapter of the Association for Computational Linguistics: Human Language Technologies",
    month = jun,
    year = "2021",
    address = "Online",
    publisher = "Association for Computational Linguistics",
    url = "https://aclanthology.org/2021.naacl-main.410/",
    doi = "10.18653/v1/2021.naacl-main.410",
    pages = "5203--5212",
}

@misc{ref:gemini2024gemini,
      title={Gemini: A Family of Highly Capable Multimodal Models}, 
      author={Gemini Team and Rohan Anil and Sebastian Borgeaud and Jean-Baptiste Alayrac and Jiahui Yu and Radu Soricut and Johan Schalkwyk and Andrew M. Dai and Anja Hauth and Katie Millican and David Silver and Melvin Johnson and Ioannis Antonoglou and Julian Schrittwieser and Amelia Glaese and Jilin Chen and Emily Pitler and Timothy Lillicrap and Angeliki Lazaridou and Orhan Firat and James Molloy and Michael Isard and Paul R. Barham and Tom Hennigan and Benjamin Lee and Fabio Viola and Malcolm Reynolds and Yuanzhong Xu and Ryan Doherty and Eli Collins and Clemens Meyer and Eliza Rutherford and Erica Moreira and Kareem Ayoub and Megha Goel and Jack Krawczyk and Cosmo Du and Ed Chi and Heng-Tze Cheng and Eric Ni and Purvi Shah and Patrick Kane and Betty Chan and Manaal Faruqui and Aliaksei Severyn and Hanzhao Lin and YaGuang Li and Yong Cheng and Abe Ittycheriah and Mahdis Mahdieh and Mia Chen and Pei Sun and Dustin Tran and Sumit Bagri and Balaji Lakshminarayanan and Jeremiah Liu and Andras Orban and Fabian Güra and Hao Zhou and Xinying Song and Aurelien Boffy and Harish Ganapathy and Steven Zheng and HyunJeong Choe and Ágoston Weisz and Tao Zhu and Yifeng Lu and Siddharth Gopal and Jarrod Kahn and Maciej Kula and Jeff Pitman and Rushin Shah and Emanuel Taropa and Majd Al Merey and Martin Baeuml and Zhifeng Chen and Laurent El Shafey and Yujing Zhang and Olcan Sercinoglu and George Tucker and Enrique Piqueras and Maxim Krikun and Iain Barr and Nikolay Savinov and Ivo Danihelka and Becca Roelofs and Anaïs White and Anders Andreassen and Tamara von Glehn and Lakshman Yagati and Mehran Kazemi and Lucas Gonzalez and Misha Khalman and Jakub Sygnowski and Alexandre Frechette and Charlotte Smith and Laura Culp and Lev Proleev and Yi Luan and Xi Chen and James Lottes and Nathan Schucher and Federico Lebron and Alban Rrustemi and Natalie Clay and Phil Crone and Tomas Kocisky and Jeffrey Zhao and Bartek Perz and Dian Yu and Heidi Howard and Adam Bloniarz and Jack W. Rae and Han Lu and Laurent Sifre and Marcello Maggioni and Fred Alcober and Dan Garrette and Megan Barnes and Shantanu Thakoor and Jacob Austin and Gabriel Barth-Maron and William Wong and Rishabh Joshi and Rahma Chaabouni and Deeni Fatiha and Arun Ahuja and Gaurav Singh Tomar and Evan Senter and Martin Chadwick and Ilya Kornakov and Nithya Attaluri and Iñaki Iturrate and Ruibo Liu and Yunxuan Li and Sarah Cogan and Jeremy Chen and Chao Jia and Chenjie Gu and Qiao Zhang and Jordan Grimstad and Ale Jakse Hartman and Xavier Garcia and Thanumalayan Sankaranarayana Pillai and Jacob Devlin and Michael Laskin and Diego de Las Casas and Dasha Valter and Connie Tao and Lorenzo Blanco and Adrià Puigdomènech Badia and David Reitter and Mianna Chen and Jenny Brennan and Clara Rivera and Sergey Brin and Shariq Iqbal and Gabriela Surita and Jane Labanowski and Abhi Rao and Stephanie Winkler and Emilio Parisotto and Yiming Gu and Kate Olszewska and Ravi Addanki and Antoine Miech and Annie Louis and Denis Teplyashin and Geoff Brown and Elliot Catt and Jan Balaguer and Jackie Xiang and Pidong Wang and Zoe Ashwood and Anton Briukhov and Albert Webson and Sanjay Ganapathy and Smit Sanghavi and Ajay Kannan and Ming-Wei Chang and Axel Stjerngren and Josip Djolonga and Yuting Sun and Ankur Bapna and Matthew Aitchison and Pedram Pejman and Henryk Michalewski and Tianhe Yu and Cindy Wang and Juliette Love and Junwhan Ahn and Dawn Bloxwich and Kehang Han and Peter Humphreys and Thibault Sellam and James Bradbury and Varun Godbole and Sina Samangooei and Bogdan Damoc and Alex Kaskasoli and Sébastien M. R. Arnold and Vijay Vasudevan and Shubham Agrawal and Jason Riesa and Dmitry Lepikhin and Richard Tanburn and Srivatsan Srinivasan and Hyeontaek Lim and Sarah Hodkinson and Pranav Shyam and Johan Ferret and Steven Hand and Ankush Garg and Tom Le Paine and Jian Li and Yujia Li and Minh Giang and Alexander Neitz and Zaheer Abbas and Sarah York and Machel Reid and Elizabeth Cole and Aakanksha Chowdhery and Dipanjan Das and Dominika Rogozińska and Vitaliy Nikolaev and Pablo Sprechmann and Zachary Nado and Lukas Zilka and Flavien Prost and Luheng He and Marianne Monteiro and Gaurav Mishra and Chris Welty and Josh Newlan and Dawei Jia and Miltiadis Allamanis and Clara Huiyi Hu and Raoul de Liedekerke and Justin Gilmer and Carl Saroufim and Shruti Rijhwani and Shaobo Hou and Disha Shrivastava and Anirudh Baddepudi and Alex Goldin and Adnan Ozturel and Albin Cassirer and Yunhan Xu and Daniel Sohn and Devendra Sachan and Reinald Kim Amplayo and Craig Swanson and Dessie Petrova and Shashi Narayan and Arthur Guez and Siddhartha Brahma and Jessica Landon and Miteyan Patel and Ruizhe Zhao and Kevin Villela and Luyu Wang and Wenhao Jia and Matthew Rahtz and Mai Giménez and Legg Yeung and James Keeling and Petko Georgiev and Diana Mincu and Boxi Wu and Salem Haykal and Rachel Saputro and Kiran Vodrahalli and James Qin and Zeynep Cankara and Abhanshu Sharma and Nick Fernando and Will Hawkins and Behnam Neyshabur and Solomon Kim and Adrian Hutter and Priyanka Agrawal and Alex Castro-Ros and George van den Driessche and Tao Wang and Fan Yang and Shuo-yiin Chang and Paul Komarek and Ross McIlroy and Mario Lučić and Guodong Zhang and Wael Farhan and Michael Sharman and Paul Natsev and Paul Michel and Yamini Bansal and Siyuan Qiao and Kris Cao and Siamak Shakeri and Christina Butterfield and Justin Chung and Paul Kishan Rubenstein and Shivani Agrawal and Arthur Mensch and Kedar Soparkar and Karel Lenc and Timothy Chung and Aedan Pope and Loren Maggiore and Jackie Kay and Priya Jhakra and Shibo Wang and Joshua Maynez and Mary Phuong and Taylor Tobin and Andrea Tacchetti and Maja Trebacz and Kevin Robinson and Yash Katariya and Sebastian Riedel and Paige Bailey and Kefan Xiao and Nimesh Ghelani and Lora Aroyo and Ambrose Slone and Neil Houlsby and Xuehan Xiong and Zhen Yang and Elena Gribovskaya and Jonas Adler and Mateo Wirth and Lisa Lee and Music Li and Thais Kagohara and Jay Pavagadhi and Sophie Bridgers and Anna Bortsova and Sanjay Ghemawat and Zafarali Ahmed and Tianqi Liu and Richard Powell and Vijay Bolina and Mariko Iinuma and Polina Zablotskaia and James Besley and Da-Woon Chung and Timothy Dozat and Ramona Comanescu and Xiance Si and Jeremy Greer and Guolong Su and Martin Polacek and Raphaël Lopez Kaufman and Simon Tokumine and Hexiang Hu and Elena Buchatskaya and Yingjie Miao and Mohamed Elhawaty and Aditya Siddhant and Nenad Tomasev and Jinwei Xing and Christina Greer and Helen Miller and Shereen Ashraf and Aurko Roy and Zizhao Zhang and Ada Ma and Angelos Filos and Milos Besta and Rory Blevins and Ted Klimenko and Chih-Kuan Yeh and Soravit Changpinyo and Jiaqi Mu and Oscar Chang and Mantas Pajarskas and Carrie Muir and Vered Cohen and Charline Le Lan and Krishna Haridasan and Amit Marathe and Steven Hansen and Sholto Douglas and Rajkumar Samuel and Mingqiu Wang and Sophia Austin and Chang Lan and Jiepu Jiang and Justin Chiu and Jaime Alonso Lorenzo and Lars Lowe Sjösund and Sébastien Cevey and Zach Gleicher and Thi Avrahami and Anudhyan Boral and Hansa Srinivasan and Vittorio Selo and Rhys May and Konstantinos Aisopos and Léonard Hussenot and Livio Baldini Soares and Kate Baumli and Michael B. Chang and Adrià Recasens and Ben Caine and Alexander Pritzel and Filip Pavetic and Fabio Pardo and Anita Gergely and Justin Frye and Vinay Ramasesh and Dan Horgan and Kartikeya Badola and Nora Kassner and Subhrajit Roy and Ethan Dyer and Víctor Campos Campos and Alex Tomala and Yunhao Tang and Dalia El Badawy and Elspeth White and Basil Mustafa and Oran Lang and Abhishek Jindal and Sharad Vikram and Zhitao Gong and Sergi Caelles and Ross Hemsley and Gregory Thornton and Fangxiaoyu Feng and Wojciech Stokowiec and Ce Zheng and Phoebe Thacker and Çağlar Ünlü and Zhishuai Zhang and Mohammad Saleh and James Svensson and Max Bileschi and Piyush Patil and Ankesh Anand and Roman Ring and Katerina Tsihlas and Arpi Vezer and Marco Selvi and Toby Shevlane and Mikel Rodriguez and Tom Kwiatkowski and Samira Daruki and Keran Rong and Allan Dafoe and Nicholas FitzGerald and Keren Gu-Lemberg and Mina Khan and Lisa Anne Hendricks and Marie Pellat and Vladimir Feinberg and James Cobon-Kerr and Tara Sainath and Maribeth Rauh and Sayed Hadi Hashemi and Richard Ives and Yana Hasson and Eric Noland and Yuan Cao and Nathan Byrd and Le Hou and Qingze Wang and Thibault Sottiaux and Michela Paganini and Jean-Baptiste Lespiau and Alexandre Moufarek and Samer Hassan and Kaushik Shivakumar and Joost van Amersfoort and Amol Mandhane and Pratik Joshi and Anirudh Goyal and Matthew Tung and Andrew Brock and Hannah Sheahan and Vedant Misra and Cheng Li and Nemanja Rakićević and Mostafa Dehghani and Fangyu Liu and Sid Mittal and Junhyuk Oh and Seb Noury and Eren Sezener and Fantine Huot and Matthew Lamm and Nicola De Cao and Charlie Chen and Sidharth Mudgal and Romina Stella and Kevin Brooks and Gautam Vasudevan and Chenxi Liu and Mainak Chain and Nivedita Melinkeri and Aaron Cohen and Venus Wang and Kristie Seymore and Sergey Zubkov and Rahul Goel and Summer Yue and Sai Krishnakumaran and Brian Albert and Nate Hurley and Motoki Sano and Anhad Mohananey and Jonah Joughin and Egor Filonov and Tomasz Kępa and Yomna Eldawy and Jiawern Lim and Rahul Rishi and Shirin Badiezadegan and Taylor Bos and Jerry Chang and Sanil Jain and Sri Gayatri Sundara Padmanabhan and Subha Puttagunta and Kalpesh Krishna and Leslie Baker and Norbert Kalb and Vamsi Bedapudi and Adam Kurzrok and Shuntong Lei and Anthony Yu and Oren Litvin and Xiang Zhou and Zhichun Wu and Sam Sobell and Andrea Siciliano and Alan Papir and Robby Neale and Jonas Bragagnolo and Tej Toor and Tina Chen and Valentin Anklin and Feiran Wang and Richie Feng and Milad Gholami and Kevin Ling and Lijuan Liu and Jules Walter and Hamid Moghaddam and Arun Kishore and Jakub Adamek and Tyler Mercado and Jonathan Mallinson and Siddhinita Wandekar and Stephen Cagle and Eran Ofek and Guillermo Garrido and Clemens Lombriser and Maksim Mukha and Botu Sun and Hafeezul Rahman Mohammad and Josip Matak and Yadi Qian and Vikas Peswani and Pawel Janus and Quan Yuan and Leif Schelin and Oana David and Ankur Garg and Yifan He and Oleksii Duzhyi and Anton Älgmyr and Timothée Lottaz and Qi Li and Vikas Yadav and Luyao Xu and Alex Chinien and Rakesh Shivanna and Aleksandr Chuklin and Josie Li and Carrie Spadine and Travis Wolfe and Kareem Mohamed and Subhabrata Das and Zihang Dai and Kyle He and Daniel von Dincklage and Shyam Upadhyay and Akanksha Maurya and Luyan Chi and Sebastian Krause and Khalid Salama and Pam G Rabinovitch and Pavan Kumar Reddy M and Aarush Selvan and Mikhail Dektiarev and Golnaz Ghiasi and Erdem Guven and Himanshu Gupta and Boyi Liu and Deepak Sharma and Idan Heimlich Shtacher and Shachi Paul and Oscar Akerlund and François-Xavier Aubet and Terry Huang and Chen Zhu and Eric Zhu and Elico Teixeira and Matthew Fritze and Francesco Bertolini and Liana-Eleonora Marinescu and Martin Bölle and Dominik Paulus and Khyatti Gupta and Tejasi Latkar and Max Chang and Jason Sanders and Roopa Wilson and Xuewei Wu and Yi-Xuan Tan and Lam Nguyen Thiet and Tulsee Doshi and Sid Lall and Swaroop Mishra and Wanming Chen and Thang Luong and Seth Benjamin and Jasmine Lee and Ewa Andrejczuk and Dominik Rabiej and Vipul Ranjan and Krzysztof Styrc and Pengcheng Yin and Jon Simon and Malcolm Rose Harriott and Mudit Bansal and Alexei Robsky and Geoff Bacon and David Greene and Daniil Mirylenka and Chen Zhou and Obaid Sarvana and Abhimanyu Goyal and Samuel Andermatt and Patrick Siegler and Ben Horn and Assaf Israel and Francesco Pongetti and Chih-Wei "Louis" Chen and Marco Selvatici and Pedro Silva and Kathie Wang and Jackson Tolins and Kelvin Guu and Roey Yogev and Xiaochen Cai and Alessandro Agostini and Maulik Shah and Hung Nguyen and Noah Ó Donnaile and Sébastien Pereira and Linda Friso and Adam Stambler and Adam Kurzrok and Chenkai Kuang and Yan Romanikhin and Mark Geller and ZJ Yan and Kane Jang and Cheng-Chun Lee and Wojciech Fica and Eric Malmi and Qijun Tan and Dan Banica and Daniel Balle and Ryan Pham and Yanping Huang and Diana Avram and Hongzhi Shi and Jasjot Singh and Chris Hidey and Niharika Ahuja and Pranab Saxena and Dan Dooley and Srividya Pranavi Potharaju and Eileen O'Neill and Anand Gokulchandran and Ryan Foley and Kai Zhao and Mike Dusenberry and Yuan Liu and Pulkit Mehta and Ragha Kotikalapudi and Chalence Safranek-Shrader and Andrew Goodman and Joshua Kessinger and Eran Globen and Prateek Kolhar and Chris Gorgolewski and Ali Ibrahim and Yang Song and Ali Eichenbaum and Thomas Brovelli and Sahitya Potluri and Preethi Lahoti and Cip Baetu and Ali Ghorbani and Charles Chen and Andy Crawford and Shalini Pal and Mukund Sridhar and Petru Gurita and Asier Mujika and Igor Petrovski and Pierre-Louis Cedoz and Chenmei Li and Shiyuan Chen and Niccolò Dal Santo and Siddharth Goyal and Jitesh Punjabi and Karthik Kappaganthu and Chester Kwak and Pallavi LV and Sarmishta Velury and Himadri Choudhury and Jamie Hall and Premal Shah and Ricardo Figueira and Matt Thomas and Minjie Lu and Ting Zhou and Chintu Kumar and Thomas Jurdi and Sharat Chikkerur and Yenai Ma and Adams Yu and Soo Kwak and Victor Ähdel and Sujeevan Rajayogam and Travis Choma and Fei Liu and Aditya Barua and Colin Ji and Ji Ho Park and Vincent Hellendoorn and Alex Bailey and Taylan Bilal and Huanjie Zhou and Mehrdad Khatir and Charles Sutton and Wojciech Rzadkowski and Fiona Macintosh and Konstantin Shagin and Paul Medina and Chen Liang and Jinjing Zhou and Pararth Shah and Yingying Bi and Attila Dankovics and Shipra Banga and Sabine Lehmann and Marissa Bredesen and Zifan Lin and John Eric Hoffmann and Jonathan Lai and Raynald Chung and Kai Yang and Nihal Balani and Arthur Bražinskas and Andrei Sozanschi and Matthew Hayes and Héctor Fernández Alcalde and Peter Makarov and Will Chen and Antonio Stella and Liselotte Snijders and Michael Mandl and Ante Kärrman and Paweł Nowak and Xinyi Wu and Alex Dyck and Krishnan Vaidyanathan and Raghavender R and Jessica Mallet and Mitch Rudominer and Eric Johnston and Sushil Mittal and Akhil Udathu and Janara Christensen and Vishal Verma and Zach Irving and Andreas Santucci and Gamaleldin Elsayed and Elnaz Davoodi and Marin Georgiev and Ian Tenney and Nan Hua and Geoffrey Cideron and Edouard Leurent and Mahmoud Alnahlawi and Ionut Georgescu and Nan Wei and Ivy Zheng and Dylan Scandinaro and Heinrich Jiang and Jasper Snoek and Mukund Sundararajan and Xuezhi Wang and Zack Ontiveros and Itay Karo and Jeremy Cole and Vinu Rajashekhar and Lara Tumeh and Eyal Ben-David and Rishub Jain and Jonathan Uesato and Romina Datta and Oskar Bunyan and Shimu Wu and John Zhang and Piotr Stanczyk and Ye Zhang and David Steiner and Subhajit Naskar and Michael Azzam and Matthew Johnson and Adam Paszke and Chung-Cheng Chiu and Jaume Sanchez Elias and Afroz Mohiuddin and Faizan Muhammad and Jin Miao and Andrew Lee and Nino Vieillard and Jane Park and Jiageng Zhang and Jeff Stanway and Drew Garmon and Abhijit Karmarkar and Zhe Dong and Jong Lee and Aviral Kumar and Luowei Zhou and Jonathan Evens and William Isaac and Geoffrey Irving and Edward Loper and Michael Fink and Isha Arkatkar and Nanxin Chen and Izhak Shafran and Ivan Petrychenko and Zhe Chen and Johnson Jia and Anselm Levskaya and Zhenkai Zhu and Peter Grabowski and Yu Mao and Alberto Magni and Kaisheng Yao and Javier Snaider and Norman Casagrande and Evan Palmer and Paul Suganthan and Alfonso Castaño and Irene Giannoumis and Wooyeol Kim and Mikołaj Rybiński and Ashwin Sreevatsa and Jennifer Prendki and David Soergel and Adrian Goedeckemeyer and Willi Gierke and Mohsen Jafari and Meenu Gaba and Jeremy Wiesner and Diana Gage Wright and Yawen Wei and Harsha Vashisht and Yana Kulizhskaya and Jay Hoover and Maigo Le and Lu Li and Chimezie Iwuanyanwu and Lu Liu and Kevin Ramirez and Andrey Khorlin and Albert Cui and Tian LIN and Marcus Wu and Ricardo Aguilar and Keith Pallo and Abhishek Chakladar and Ginger Perng and Elena Allica Abellan and Mingyang Zhang and Ishita Dasgupta and Nate Kushman and Ivo Penchev and Alena Repina and Xihui Wu and Tom van der Weide and Priya Ponnapalli and Caroline Kaplan and Jiri Simsa and Shuangfeng Li and Olivier Dousse and Fan Yang and Jeff Piper and Nathan Ie and Rama Pasumarthi and Nathan Lintz and Anitha Vijayakumar and Daniel Andor and Pedro Valenzuela and Minnie Lui and Cosmin Paduraru and Daiyi Peng and Katherine Lee and Shuyuan Zhang and Somer Greene and Duc Dung Nguyen and Paula Kurylowicz and Cassidy Hardin and Lucas Dixon and Lili Janzer and Kiam Choo and Ziqiang Feng and Biao Zhang and Achintya Singhal and Dayou Du and Dan McKinnon and Natasha Antropova and Tolga Bolukbasi and Orgad Keller and David Reid and Daniel Finchelstein and Maria Abi Raad and Remi Crocker and Peter Hawkins and Robert Dadashi and Colin Gaffney and Ken Franko and Anna Bulanova and Rémi Leblond and Shirley Chung and Harry Askham and Luis C. Cobo and Kelvin Xu and Felix Fischer and Jun Xu and Christina Sorokin and Chris Alberti and Chu-Cheng Lin and Colin Evans and Alek Dimitriev and Hannah Forbes and Dylan Banarse and Zora Tung and Mark Omernick and Colton Bishop and Rachel Sterneck and Rohan Jain and Jiawei Xia and Ehsan Amid and Francesco Piccinno and Xingyu Wang and Praseem Banzal and Daniel J. Mankowitz and Alex Polozov and Victoria Krakovna and Sasha Brown and MohammadHossein Bateni and Dennis Duan and Vlad Firoiu and Meghana Thotakuri and Tom Natan and Matthieu Geist and Ser tan Girgin and Hui Li and Jiayu Ye and Ofir Roval and Reiko Tojo and Michael Kwong and James Lee-Thorp and Christopher Yew and Danila Sinopalnikov and Sabela Ramos and John Mellor and Abhishek Sharma and Kathy Wu and David Miller and Nicolas Sonnerat and Denis Vnukov and Rory Greig and Jennifer Beattie and Emily Caveness and Libin Bai and Julian Eisenschlos and Alex Korchemniy and Tomy Tsai and Mimi Jasarevic and Weize Kong and Phuong Dao and Zeyu Zheng and Frederick Liu and Fan Yang and Rui Zhu and Tian Huey Teh and Jason Sanmiya and Evgeny Gladchenko and Nejc Trdin and Daniel Toyama and Evan Rosen and Sasan Tavakkol and Linting Xue and Chen Elkind and Oliver Woodman and John Carpenter and George Papamakarios and Rupert Kemp and Sushant Kafle and Tanya Grunina and Rishika Sinha and Alice Talbert and Diane Wu and Denese Owusu-Afriyie and Cosmo Du and Chloe Thornton and Jordi Pont-Tuset and Pradyumna Narayana and Jing Li and Saaber Fatehi and John Wieting and Omar Ajmeri and Benigno Uria and Yeongil Ko and Laura Knight and Amélie Héliou and Ning Niu and Shane Gu and Chenxi Pang and Yeqing Li and Nir Levine and Ariel Stolovich and Rebeca Santamaria-Fernandez and Sonam Goenka and Wenny Yustalim and Robin Strudel and Ali Elqursh and Charlie Deck and Hyo Lee and Zonglin Li and Kyle Levin and Raphael Hoffmann and Dan Holtmann-Rice and Olivier Bachem and Sho Arora and Christy Koh and Soheil Hassas Yeganeh and Siim Põder and Mukarram Tariq and Yanhua Sun and Lucian Ionita and Mojtaba Seyedhosseini and Pouya Tafti and Zhiyu Liu and Anmol Gulati and Jasmine Liu and Xinyu Ye and Bart Chrzaszcz and Lily Wang and Nikhil Sethi and Tianrun Li and Ben Brown and Shreya Singh and Wei Fan and Aaron Parisi and Joe Stanton and Vinod Koverkathu and Christopher A. Choquette-Choo and Yunjie Li and TJ Lu and Abe Ittycheriah and Prakash Shroff and Mani Varadarajan and Sanaz Bahargam and Rob Willoughby and David Gaddy and Guillaume Desjardins and Marco Cornero and Brona Robenek and Bhavishya Mittal and Ben Albrecht and Ashish Shenoy and Fedor Moiseev and Henrik Jacobsson and Alireza Ghaffarkhah and Morgane Rivière and Alanna Walton and Clément Crepy and Alicia Parrish and Zongwei Zhou and Clement Farabet and Carey Radebaugh and Praveen Srinivasan and Claudia van der Salm and Andreas Fidjeland and Salvatore Scellato and Eri Latorre-Chimoto and Hanna Klimczak-Plucińska and David Bridson and Dario de Cesare and Tom Hudson and Piermaria Mendolicchio and Lexi Walker and Alex Morris and Matthew Mauger and Alexey Guseynov and Alison Reid and Seth Odoom and Lucia Loher and Victor Cotruta and Madhavi Yenugula and Dominik Grewe and Anastasia Petrushkina and Tom Duerig and Antonio Sanchez and Steve Yadlowsky and Amy Shen and Amir Globerson and Lynette Webb and Sahil Dua and Dong Li and Surya Bhupatiraju and Dan Hurt and Haroon Qureshi and Ananth Agarwal and Tomer Shani and Matan Eyal and Anuj Khare and Shreyas Rammohan Belle and Lei Wang and Chetan Tekur and Mihir Sanjay Kale and Jinliang Wei and Ruoxin Sang and Brennan Saeta and Tyler Liechty and Yi Sun and Yao Zhao and Stephan Lee and Pandu Nayak and Doug Fritz and Manish Reddy Vuyyuru and John Aslanides and Nidhi Vyas and Martin Wicke and Xiao Ma and Evgenii Eltyshev and Nina Martin and Hardie Cate and James Manyika and Keyvan Amiri and Yelin Kim and Xi Xiong and Kai Kang and Florian Luisier and Nilesh Tripuraneni and David Madras and Mandy Guo and Austin Waters and Oliver Wang and Joshua Ainslie and Jason Baldridge and Han Zhang and Garima Pruthi and Jakob Bauer and Feng Yang and Riham Mansour and Jason Gelman and Yang Xu and George Polovets and Ji Liu and Honglong Cai and Warren Chen and XiangHai Sheng and Emily Xue and Sherjil Ozair and Christof Angermueller and Xiaowei Li and Anoop Sinha and Weiren Wang and Julia Wiesinger and Emmanouil Koukoumidis and Yuan Tian and Anand Iyer and Madhu Gurumurthy and Mark Goldenson and Parashar Shah and MK Blake and Hongkun Yu and Anthony Urbanowicz and Jennimaria Palomaki and Chrisantha Fernando and Ken Durden and Harsh Mehta and Nikola Momchev and Elahe Rahimtoroghi and Maria Georgaki and Amit Raul and Sebastian Ruder and Morgan Redshaw and Jinhyuk Lee and Denny Zhou and Komal Jalan and Dinghua Li and Blake Hechtman and Parker Schuh and Milad Nasr and Kieran Milan and Vladimir Mikulik and Juliana Franco and Tim Green and Nam Nguyen and Joe Kelley and Aroma Mahendru and Andrea Hu and Joshua Howland and Ben Vargas and Jeffrey Hui and Kshitij Bansal and Vikram Rao and Rakesh Ghiya and Emma Wang and Ke Ye and Jean Michel Sarr and Melanie Moranski Preston and Madeleine Elish and Steve Li and Aakash Kaku and Jigar Gupta and Ice Pasupat and Da-Cheng Juan and Milan Someswar and Tejvi M. and Xinyun Chen and Aida Amini and Alex Fabrikant and Eric Chu and Xuanyi Dong and Amruta Muthal and Senaka Buthpitiya and Sarthak Jauhari and Nan Hua and Urvashi Khandelwal and Ayal Hitron and Jie Ren and Larissa Rinaldi and Shahar Drath and Avigail Dabush and Nan-Jiang Jiang and Harshal Godhia and Uli Sachs and Anthony Chen and Yicheng Fan and Hagai Taitelbaum and Hila Noga and Zhuyun Dai and James Wang and Chen Liang and Jenny Hamer and Chun-Sung Ferng and Chenel Elkind and Aviel Atias and Paulina Lee and Vít Listík and Mathias Carlen and Jan van de Kerkhof and Marcin Pikus and Krunoslav Zaher and Paul Müller and Sasha Zykova and Richard Stefanec and Vitaly Gatsko and Christoph Hirnschall and Ashwin Sethi and Xingyu Federico Xu and Chetan Ahuja and Beth Tsai and Anca Stefanoiu and Bo Feng and Keshav Dhandhania and Manish Katyal and Akshay Gupta and Atharva Parulekar and Divya Pitta and Jing Zhao and Vivaan Bhatia and Yashodha Bhavnani and Omar Alhadlaq and Xiaolin Li and Peter Danenberg and Dennis Tu and Alex Pine and Vera Filippova and Abhipso Ghosh and Ben Limonchik and Bhargava Urala and Chaitanya Krishna Lanka and Derik Clive and Yi Sun and Edward Li and Hao Wu and Kevin Hongtongsak and Ianna Li and Kalind Thakkar and Kuanysh Omarov and Kushal Majmundar and Michael Alverson and Michael Kucharski and Mohak Patel and Mudit Jain and Maksim Zabelin and Paolo Pelagatti and Rohan Kohli and Saurabh Kumar and Joseph Kim and Swetha Sankar and Vineet Shah and Lakshmi Ramachandruni and Xiangkai Zeng and Ben Bariach and Laura Weidinger and Tu Vu and Alek Andreev and Antoine He and Kevin Hui and Sheleem Kashem and Amar Subramanya and Sissie Hsiao and Demis Hassabis and Koray Kavukcuoglu and Adam Sadovsky and Quoc Le and Trevor Strohman and Yonghui Wu and Slav Petrov and Jeffrey Dean and Oriol Vinyals},
      year={2024},
      eprint={2312.11805},
      archivePrefix={arXiv},
      primaryClass={cs.CL},
      url={https://arxiv.org/abs/2312.11805}, 
}

@inproceedings{ref:bayat2024enhanced,
  title={Enhanced language model truthfulness with learnable intervention and uncertainty expression},
  author={Bayat, Farima Fatahi and Liu, Xin and Jagadish, H and Wang, Lu},
  booktitle={Findings of the Association for Computational Linguistics ACL 2024},
  pages={12388--12400},
  year={2024}
}

@article{ref:nguyen2025risk,
  title={Risk-aware distributional intervention policies for language models},
  author={Nguyen, Bao and Nguyen, Binh and Nguyen, Duy and Nguyen, Viet Anh},
  journal={arXiv preprint arXiv:2501.15758},
  year={2025}
}

@inproceedings{ref:parrish2022bbq,
    title = "{BBQ}: A hand-built bias benchmark for question answering",
    author = "Parrish, Alicia  and
      Chen, Angelica  and
      Nangia, Nikita  and
      Padmakumar, Vishakh  and
      Phang, Jason  and
      Thompson, Jana  and
      Htut, Phu Mon  and
      Bowman, Samuel",
    editor = "Muresan, Smaranda  and
      Nakov, Preslav  and
      Villavicencio, Aline",
    booktitle = "Findings of the Association for Computational Linguistics: ACL 2022",
    month = may,
    year = "2022",
    address = "Dublin, Ireland",
    publisher = "Association for Computational Linguistics",
    url = "https://aclanthology.org/2022.findings-acl.165/",
    doi = "10.18653/v1/2022.findings-acl.165",
    pages = "2086--2105",
}

@inproceedings{ref:logacheva2022paradetox,
    title = "{P}ara{D}etox: Detoxification with Parallel Data",
    author = "Logacheva, Varvara  and
      Dementieva, Daryna  and
      Ustyantsev, Sergey  and
      Moskovskiy, Daniil  and
      Dale, David  and
      Krotova, Irina  and
      Semenov, Nikita  and
      Panchenko, Alexander",
    editor = "Muresan, Smaranda  and
      Nakov, Preslav  and
      Villavicencio, Aline",
    booktitle = "Proceedings of the 60th Annual Meeting of the Association for Computational Linguistics (Volume 1: Long Papers)",
    month = may,
    year = "2022",
    address = "Dublin, Ireland",
    publisher = "Association for Computational Linguistics",
    url = "https://aclanthology.org/2022.acl-long.469/",
    doi = "10.18653/v1/2022.acl-long.469",
    pages = "6804--6818",
}

@misc{ref:bai2022training,
      title={Training a Helpful and Harmless Assistant with Reinforcement Learning from Human Feedback}, 
      author={Yuntao Bai and Andy Jones and Kamal Ndousse and Amanda Askell and Anna Chen and Nova DasSarma and Dawn Drain and Stanislav Fort and Deep Ganguli and Tom Henighan and Nicholas Joseph and Saurav Kadavath and Jackson Kernion and Tom Conerly and Sheer El-Showk and Nelson Elhage and Zac Hatfield-Dodds and Danny Hernandez and Tristan Hume and Scott Johnston and Shauna Kravec and Liane Lovitt and Neel Nanda and Catherine Olsson and Dario Amodei and Tom Brown and Jack Clark and Sam McCandlish and Chris Olah and Ben Mann and Jared Kaplan},
      year={2022},
      eprint={2204.05862},
      archivePrefix={arXiv},
      primaryClass={cs.CL},
      url={https://arxiv.org/abs/2204.05862}, 
}

@inproceedings{ref:OpenBookQA2018,
 title={Can a Suit of Armor Conduct Electricity? A New Dataset for Open Book Question Answering},
 author={Todor Mihaylov and Peter Clark and Tushar Khot and Ashish Sabharwal},
 booktitle={EMNLP},
 year={2018}
}

@article{ref:jiang2024mixtral,
  title={Mixtral of experts},
  author={Jiang, Albert Q and Sablayrolles, Alexandre and Roux, Antoine and Mensch, Arthur and Savary, Blanche and Bamford, Chris and Chaplot, Devendra Singh and Casas, Diego de las and Hanna, Emma Bou and Bressand, Florian and others},
  journal={arXiv preprint arXiv:2401.04088},
  year={2024}
}

@inproceedings{nguyen2025laser,
    title={{LAS}eR: Learning to Adaptively Select Reward Models with Multi-Arm Bandits},
    author={Duy Nguyen and Archiki Prasad and Elias Stengel-Eskin and Mohit Bansal},
    booktitle={The Thirty-ninth Annual Conference on Neural Information Processing Systems},
    year={2025},
    url={https://openreview.net/forum?id=tSpWkTFASC}
}

\appendix

\section{Experimental Settings} \label{sec:app-setting}
\paragraph{Data Preprocessing.}  For the TruthfulQA dataset (Apache-2.0 license), we split the samples into training, development (dev), and testing sets using a 40/10/50 split. For Toxigen (MIT license) and BBQ (cc-by-4.0 license), these datasets have already been split into training and validation sets. We use the validation sets to test the models while further splitting the training sets with an 80/20 ratio to create new train and dev sets. For HelpSteer (released under cc-by-4.0 license), after sampling 500 positive and 500 negative samples for each attribute, we apply a 40/10/50 split for the train-dev-test sets.
All methods are trained on the combined training sets from different datasets and evaluated on the corresponding test sets for each task individually.
All datasets are in English.

\paragraph{Implementation Details for Baselines and \method{}.} We provide implementation details of our method and baselines as follows:

\begin{itemize}
    \item \textbf{Layer to intervene:} Following previous work~\citep{ref:li2024inference}, which suggests that information is primarily processed in the early to middle layers, we conduct a grid search from layer 10 to layer 22 of LLMs to maximize performance on the dev set. 
    Based on this search, we intervene at layer 14 for Llama-3.1-8B and Llama-3.1-8B Chat and at layer 16 for Qwen2.5-7B.
    \item \textbf{Training:} For training with LoRA, we set the rank to $16$ and alpha to $32$. We fine-tune the model for $10$ iterations using a learning rate of $5e\!-\!6$ and a batch size of $16$. For our method, we use a batch size of $96$ for QA tasks and $160$ for generation tasks, while each batch contains $16$ positive and $16$ negative samples for each attribute.
    \item \textbf{Hyperparameters:} For intervention baselines, we follow the same settings as in the original paper for TruthfulQA. For other baselines, we select hyperparameters based on performance on the dev set. For \method{}, we set $\lambda_{\mathrm{pos}} = \lambda_{\mathrm{sparse}}$, as we assume that the weights of constraints applied to positive and negative samples should be the same. We then perform a grid search on the dev set for $\lambda_{\mathrm{pos}}$, $\lambda_{\mathrm{sparse}}$, and $\lambda_{\mathrm{ortho}}$ in the range $[0, 1]$ with a step size of 0.1.  For QA tasks, the optimal hyperparameters are $\lambda_{\mathrm{pos}} = \lambda_{\mathrm{sparse}} = 0.9$ and $\lambda_{\mathrm{ortho}} = 0.1$. For generation tasks, the optimal hyperparameters are $\lambda_{\mathrm{pos}} = \lambda_{\mathrm{sparse}} = 0.8$ and $\lambda_{\mathrm{ortho}} = 0.1$.  
\end{itemize}

\paragraph{MMD Kernel.} The MMD loss in~\eqref{eq:mmd-sum} can also be written as:
\begin{align}
\mc{L}_{\mathrm{MMD}} = \sum_{t=1}^{T} \Bigg( 
    & \frac{1}{|\mc A_t^\mathbf{p}|^2}  
      \sum_{\substack{a_i, a_j \in \mc A_t^\mathbf{p}}} k(a_i, a_j) \notag \\
    & + \frac{1}{|\mc A_t^\mathbf{n}|^2}  
      \sum_{\substack{a_i, a_j \in \mc A_t^\mathbf{n}}} k(f(a_i), f(a_j)) \notag \\
    & - \frac{2}{|\mc A_t^\mathbf{p}||\mc A_t^\mathbf{n}|}  
      \sum_{\substack{a_i \in \mc A_t^\mathbf{p} \\ a_j \in \mc A_t^\mathbf{n}}} k(a_i, f(a_j))
\Bigg).
\end{align}

In our experiments, the MMD loss is computed using a Gaussian kernel:
\[
k(x,y) = \exp\left(-\frac{\|x-y\|^2}{2\sigma^2}\right),
\]
with a carefully chosen bandwidth $\sigma$ based on the performance on the dev set (minimizes the validation loss). We choose $\sigma=2$ for both QA and generation tasks.

\paragraph{GPUs.} All of our experiments are run on four RTX A6000 with 48G memory each.

\paragraph{Prompts.} For TruthfulQA, Toxigen, BBQ, and HelpSteer, we do not include any system instruction prompt. Instead, we simply provide the input questions and capture the responses. For FaithEval, we follow the same prompt design as described in the original paper~\citep{ref:ming2024faitheval}.

\section{Further Analysis of \method{}} \label{sec:analysis}

\begin{figure}[t]
    \centering
    \includegraphics[width=1.0\linewidth]{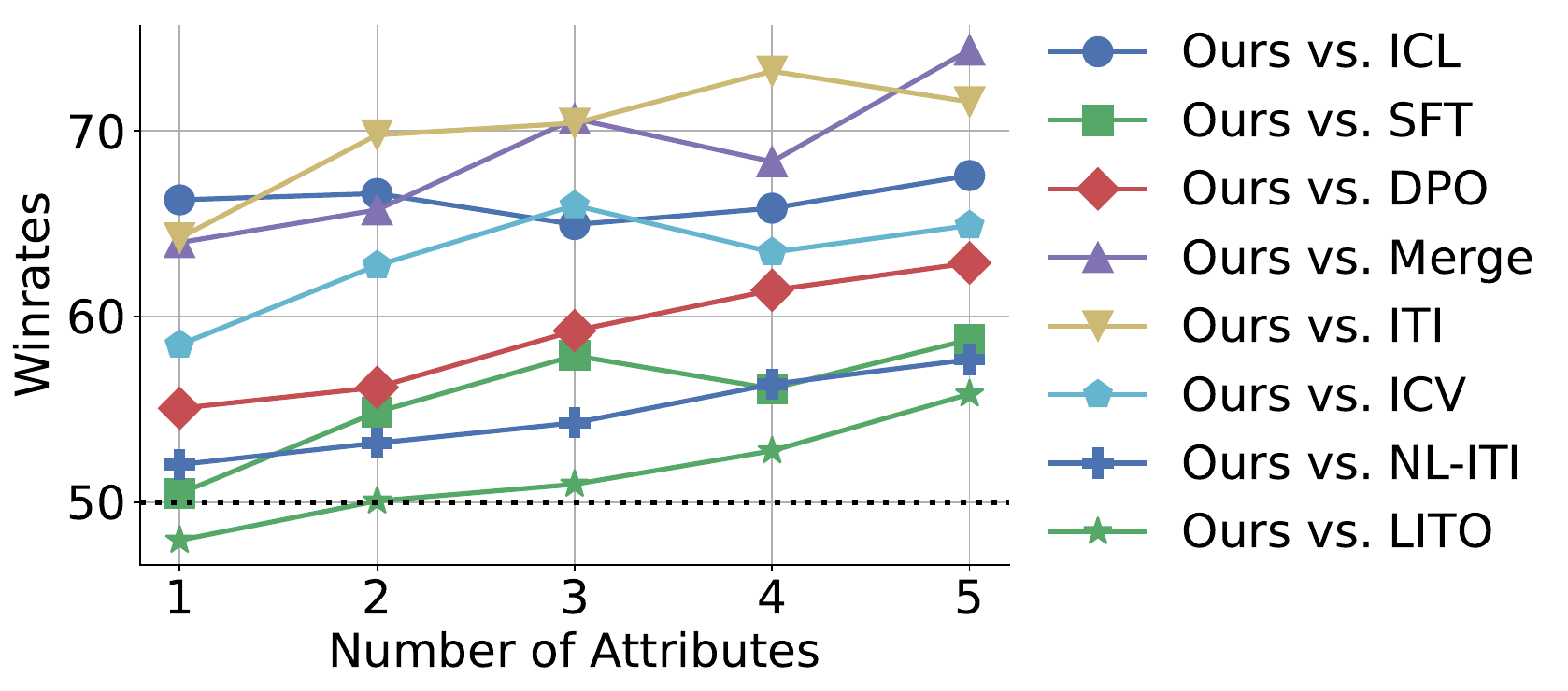}
    \caption{Win rates of \method{} vs. baselines as the number of HelpSteer attributes increases from 1 to 5. The dotted line represents a $50\%$ win-rate, indicating a tie.}
    \label{fig:scaling}
\end{figure}

\paragraph{\method{} Performs Better when Scaling the Number of Attributes for Generation.} In real-world applications, it is often the case that objectives must be adopted sequentially (rather than having all objectives presented simultaneously). 
\cref{fig:scaling} illustrates how our method scales as we increase the number of attributes in HelpSteer from 1 to 5, reporting the win rates against various baselines. Each point on the horizontal axis corresponds to a scenario where the model must optimize for a specific number of attributes (e.g., helpfulness, correctness, and coherence). 
We observe that our approach maintains or even improves its margin over baselines as the number of attributes grows. While these baselines typically experience diminishing outcomes or exhibit flat performance when handling multiple objectives, \method{} effectively balances those objectives. For example, when scaling from 1 to 4 attributes, \method{} increases the win rate margin over SFT by 5.65\% and over ITI by 8.98\%. This result highlights the robustness of our method in multi-attribute settings, where aligning model outputs with multiple human-preferred criteria becomes more challenging.

\paragraph{Token Selection Analysis.} We further investigate the role of token selection in our intervention framework. 
We compare several selection baselines: uniform intervention on all tokens, intervention on only the last token in the prompt, random token selection, and \method{}'s selection method. 
Table~\ref{tab:token-seletion} reports the multiple-choice accuracy on the TruthfulQA, Toxigen, and BBQ datasets. 
\method{} outperforms all other methods, highlighting that selecting the \emph{right} tokens -- rather than merely reducing the number of tokens -- plays a crucial role in achieving superior attribute-specific improvements while preserving essential contextual information.

\begin{table}[t]
\small
\centering
\begin{tabular}{lccc}
\toprule
\textbf{Method}   & \textbf{TruthfulQA} & \textbf{Toxigen} & \textbf{BBQ} \\ \midrule
Base Model        & 54.06               & 52.27            & 56.71        \\
Random            & 54.37               & 53.63            & 55.24        \\
Last               & 56.14               & 52.38            & 57.04        \\
All              & 58.09               & 55.92            & 59.74        \\
\method{}              & \textbf{61.94}      & \textbf{57.59}   & \textbf{60.32} \\ 
\bottomrule
\end{tabular}
\caption{Performance comparison of different token selection methods across three datasets on Llama-3.1-8B. 
\method{} outperforms all other strategies across all datasets, with the highest results highlighted in bold.}
\label{tab:token-seletion}
\end{table}

\section{Additional Numerical Results}\label{sec:app-results}
\paragraph{Additional Results on Other Tasks.} \modified{Since our main experiments focus on QA tasks targeting multiple trustworthiness-related attributes, we have extended our evaluation to the HH-RLHF benchmark~\citep{ref:bai2022training} to assess the generalization of MAT-Steer and baselines. In particular, we use the pretrained vectors from QA tasks in~\cref{tab:qa-tasks} and test them on the HH-RLHF test set. The HH-RLHF benchmark involves complex, real-world assistant-style queries, covering open-ended topics that require models to be both helpful and harmless. We use GPT-4o to compare generations from MAT-Steer and three baselines: SFT, ICL, and LITO (the strongest baseline in all benchmarks) on the HH-RLHF test set. The GPT-4o win rates are shown in~\cref{tab:hh-rlhf}. These results show that MAT-Steer outperforms all baselines, demonstrating its ability to generalize beyond in-domain QA tasks.}

\begin{table}[h]
\centering
\begin{tabular}{lc}
\toprule
\textbf{Comparison} & \textbf{GPT-4o Win Rates} \\
\midrule
\method{} vs. ICL  & 67.07\% \\
\method{} vs. SFT  & 62.35\% \\
\method{} vs. LITO & 59.12\% \\
\bottomrule
\end{tabular}
\caption{GPT-4o win rates comparing MAT-Steer against other methods on HH-RLHF.}
\label{tab:hh-rlhf}
\end{table}

\modified{Additionally, we extend MAT-Steer to an explicitly structured reasoning task by incorporating the OpenBookQA (OBQA)~\citep{ref:OpenBookQA2018} dataset into our QA tasks (alongside TruthfulQA, Toxigen, and BBQ). OBQA has been used in prior ITI work, such as NL-ITI~\citep{ref:hoscilowicz2024non} and multiple efforts on LLM reasoning~\citep{ref:dubey2024llama, ref:jiang2024mixtral}. We compare MAT-Steer on OBQA against three baselines: ICL, SFT (fine-tune the model checkpoint from 3 datasets with additional OBQA training data), and LITO (the strongest ITI baseline in all benchmarks).}

\modified{As suggested by~\cref{fig:scaling}, most baselines degrade when the number of tasks or attributes increases; therefore, we report both the accuracy on OBQA and the average performance drop on the original QA datasets (TruthfulQA, Toxigen, and BBQ) to assess multi-task robustness.~\cref{tab:obqa} demonstrates that while SFT performs slightly better than MAT-Steer on OBQA, it has a significant performance drop of 2.74\% on the other tasks. Additionally, ICL has the smallest performance drop (0.14\%) but also has the lowest OBQA accuracy (73.40\%), indicating limited overall capability. In contrast, MAT-Steer maintains strong performance across all tasks with only 0.32\% drop, highlighting its ability to generalize to new reasoning settings while preserving previously aligned behaviors, a key advantage of our multi-attribute steering framework.}

\begin{table}[h!]
\small
\centering
\begin{tabular}{lcc}
\toprule
\textbf{Method} & \textbf{OBQA (\%)} & \textbf{Avg. decrease (\%)} \\
\midrule
ICL        & 73.40 & -0.14 \\
SFT        & 77.92 & -2.74 \\
LITO       & 74.57 & -1.12 \\
\method{}  & 77.46 & -0.32 \\
\bottomrule
\end{tabular}
\caption{Performance on OBQA and average decrease on 3 original QA tasks for different methods.}
\label{tab:obqa}
\end{table}

\paragraph{Generalization to Model Type and Family.}  
We evaluate our method on different LLM families to demonstrate its robustness. Experiments on Llama-3.1-7B Chat~\citep{ref:dubey2024llama} and Qwen2.5-7B~\citep{ref:qwen2024moe} reveal that our intervention strategy transfers effectively across model architectures, providing consistent improvements in multiple-choice accuracy and generation quality. This suggests that the benefits of our approach are not limited to a single model but generalize across various base LLMs.

\begin{table}[h]
\centering
\small
\begin{tabular}{lccc}
\toprule
\textbf{Method}   & \textbf{TruthfulQA} & \textbf{Toxigen} & \textbf{BBQ} \\ \midrule
Qwen2.5-7B        & 54.06               & 52.27            & 56.71        \\ \midrule
ICL & 57.12               & 55.65            & 58.26        \\ \midrule
SFT              & 56.52               & 57.76            & 60.78        \\
DPO              & 59.56               & 55.45            & 60.43        \\ \midrule
Merge & 57.16               & 56.02            & 58.39        \\ 
RAdapt            &    57.50           &   54.87          & 58.67        \\ \midrule
ITI              &   58.09               & 58.15            & 59.74      \\
ICV              & 59.94               & 56.76            & 59.89        \\
NL-ITI          & 61.45               & 57.56            & 60.01        \\
LITO             & 62.37               & 58.29            & 60.34        \\ \midrule
\method{} \textit{(Ours)}             & \textbf{64.36}      & \textbf{60.41}   & \textbf{62.59} \\ 
\bottomrule
\end{tabular}
\caption{Performance comparison of \method{} against in-context learning, fine-tuning, multiple adapters, and intervention methods on Qwen2.5-7B.  }
\label{tab:qa-tasks-qwen}
\end{table}

\begin{table}[h]
\centering
\small
\begin{tabular}{lccc}
\toprule
\textbf{Method}   & \textbf{TruthfulQA} & \textbf{Toxigen} & \textbf{BBQ} \\ \midrule
Llama-3.1-Chat       & 51.20               & 49.97            & 53.13        \\ \midrule
ICL              & 54.47               & 55.37            & 57.24        \\ \midrule
SFT              & 58.26               & 56.45            & 59.65        \\
DPO              & 56.83               & 56.11            & 57.37        \\ \midrule
Merge            & 53.32               & 54.06            & 55.42        \\ 
RAdapt            & 57.91              & 54.74            & 56.89        \\ \midrule
ITI              & 52.57               & 52.06            & 54.56      \\
ICV              & 56.41               & 53.03            & 56.94        \\
NL-ITI           & 54.16               & 52.37            & 53.98        \\
LITO             & 61.29               & 56.94   &  58.63       \\ \midrule
\method{} \textit{(Ours)}             & \textbf{62.42}      & \textbf{57.82}            & \textbf{61.25} \\ 
\bottomrule
\end{tabular}
\caption{Performance comparison of \method{} against in-context learning, fine-tuning, multiple adapters, and intervention methods on the Llama-3.1-8B-Chat model.}
\label{tab:qa-tasks-chat}
\end{table}

\paragraph{Integration with ICL and Fine-Tuning.}

We assess the complementarity of our method when combined with other adaptation techniques, such as in-context learning and fine-tuning. We incorporate our token-level intervention strategy on top of few-shot prompting and LoRA-based fine-tuning. As reported in Table~\ref{tab:fsp-iti}, our approach further enhances the performance of both in-context learning (ICL) and fine-tuning (SFT, DPO), yielding higher accuracies on QA tasks. These results confirm that our method is not only effective as a standalone intervention strategy but also synergizes well with existing techniques to boost overall performance.

\begin{table}[t]
\centering
\small
\setlength{\tabcolsep}{2 pt}
\begin{tabular}{lcccr}
\toprule
\textbf{Method}   & \textbf{TruthfulQA} & \textbf{Toxigen} & \textbf{BBQ} \\ \midrule
ICL & 55.32               &   51.26          &   56.46      \\
ICL + \method{} &    \textbf{62.66}            &   \textbf{58.34}          &   \textbf{63.49}      \\ \midrule
SFT              & 54.02               &   55.51          &     57.29    \\
SFT + \method{}   & \textbf{61.50}          &       \textbf{62.83}         &   \textbf{64.41}          &         \\ \midrule
DPO              & 56.10              &    55.94         &    57.51     \\
DPO + \method{}  &   \textbf{63.53}         &    \textbf{63.09}            &   \textbf{64.57}          &         \\ 
\bottomrule
\end{tabular}
\caption{Performance comparison of our method against few-shot prompting, fine-tuning method with our method on top. Each method is evaluated on three datasets: TruthfulQA, Toxigen, and BBQ. The highest performance for each dataset is highlighted in bold.}
\label{tab:fsp-iti}
\end{table}

\paragraph{Examples on FaithEval.} To better understand the interpretability of our proposed gating function, we analyze the intervention magnitude of individual tokens, identifying the top $5$ tokens with the highest overall intervention magnitude. Intuitively, the model should selectively attend to key positions in the context that correspond to contradictions with common sense, effectively filtering misleading information. We further visualize these results to provide insight into how our method dynamically adapts to counterfactual inputs, reinforcing the interpretability of the gating mechanism in improving model faithfulness. For example, in Example 1, the model highlights key tokens like \emph{``water,'' ``physiological,''} and \emph{``management''} to counter the misleading context. By focusing on these tokens, the intervened model correctly determines that leaves grow at the top of trees to \emph{``capture sunlight''} rather than to \emph{``collect water''}. 

\begin{tcolorbox} [colback=gray!10, colframe=black, title=FaithEval Example 1, label=example1]
C: When \dots to optimize \highlight{water} collection and retention. This mechanism ensures that even in drier periods, trees can sustain their essential \highlight{physiological} processes through proficient \highlight{water} \highlight{management} strategies typically occurring at the tree's apex.

Q: Why do most of the \highlight{leaves} of forest trees grow at the top of the tree?

\textbf{Wrong answer:} to collect water \\
\textbf{Correct answer:} to capture sunlight
\end{tcolorbox}

\begin{tcolorbox}[colback=gray!10, colframe=black, title=FaithEval Example 2, label=example2]
C: "...One intriguing property of \highlight{wood} that has often been overlooked is its \highlight{magnetic} nature... These findings pointed to the presence of iron-like compounds within the \highlight{cellular} structure of wood, which could exhibit faint magnetic properties... early shipbuilders used \highlight{magnetized} wood..."

Q: Which statement best explains why a tree branch floats on water?

\textbf{Wrong answer:} Wood is buoyant \\
\textbf{Correct answer:} Wood is magnetic
\end{tcolorbox}

\begin{tcolorbox}[colback=gray!10, colframe=black, title=FaithEval Example 3, label=example3]
C: Understanding... yet highly effective piece of \highlight{equipment} that has been indispensable is the compass... Historical records indicate that early American exploration teams extensively \highlight{relied} on compasses to map large swathes of unexplored terrain accurately... resulting in detailed topographical maps that are still highly \highlight{valued} today for their finesse and precision in \highlight{capturing} the nuances of the landscape. These detailed maps were often cross-verified with astronomical \highlight{observations}, which reinforced the accuracy of the compass readings.

Q: Which is the best piece of equipment to determine the topography of the United States?

\textbf{Wrong answer:} compass \\
\textbf{Correct answer:} satellite
\end{tcolorbox}

\end{document}